\pdfoutput=1
\documentclass[11pt,a4paper]{article}
\usepackage[T1]{fontenc}
\usepackage[utf8]{inputenc}
\usepackage{lmodern}
\usepackage[a4paper,margin=1in]{geometry}
\usepackage{setspace}
\usepackage{authblk}
\usepackage{amsmath}
\usepackage{amssymb}
\usepackage{amsfonts}
\usepackage{mathtools}
\usepackage{bm}
\usepackage{graphicx}
\usepackage{float}
\usepackage{subcaption}
\usepackage{booktabs}
\usepackage{multirow}
\usepackage{array}
\usepackage{siunitx}
\graphicspath{{Figures/}}
\usepackage{algorithm}
\usepackage{algpseudocode}
\usepackage[numbers,sort&compress]{natbib}
\usepackage{xcolor}
\usepackage[
    colorlinks=true,
    linkcolor=blue,
    citecolor=blue,
    urlcolor=blue
]{hyperref}
\usepackage[nameinlink,noabbrev]{cleveref}
\begin{document}
\title{\textbf{Kinematic analysis of a double-closed-loop four-bar chain-based leg for a multimodal quadruped-pentarotor robot}}

\author[]{Prashant Raturi}
\author[]{Santhakumar Mohan\thanks{Corresponding author, Email: santhakumar@iitpkd.ac.in}}

\affil[]{Department of Mechanical Engineering and Center for Robotics and Control, Indian Institute of Technology Palakkad, Kerala, India}
\date{}
\maketitle
\begin{abstract}
This paper presents the kinematic modeling and singularity analysis of a three-degree-of-freedom, double four-bar closed-loop quadruped leg mechanism. This leg is essential for transitioning between the two operational modes; therefore, it is examined in both joint and task spaces, leading to the development of its Jacobian matrix, which is vital for designing the gaits required for terrestrial mode and mid-air aerial reconfiguration (for aerial mode). Various gaits are devised for the system's motion in the MATLAB simulation, and their practical implementation on the single-leg test rig is demonstrated.
\end{abstract}

\section{Introduction}
\begin{figure}
    \centering
    \includegraphics[width=\linewidth]{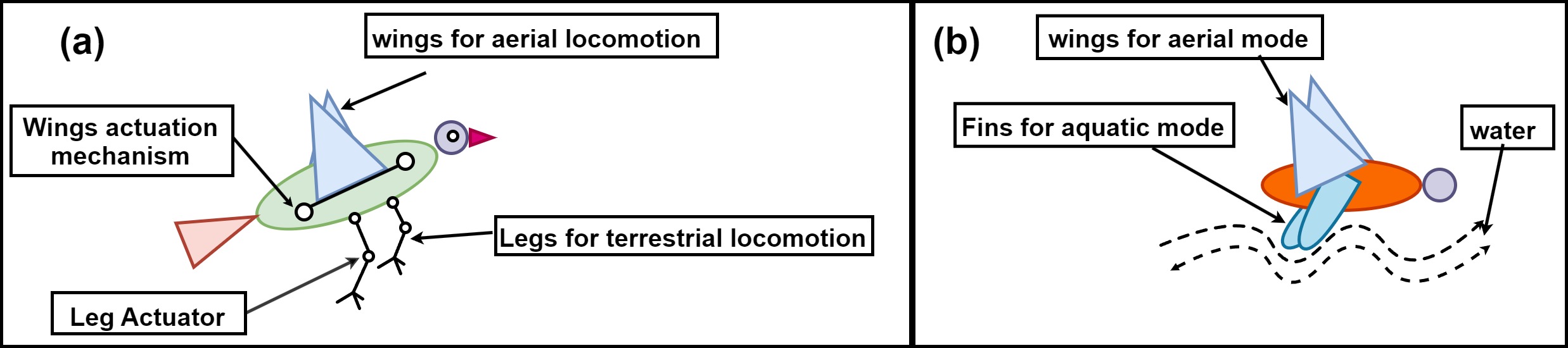}
    \caption{(a) Ground and aerial bio-inspired robot (b) Aerial and aquatic bio-inspired robot}
    \label{bio_inspired_robot}
\end{figure}

\begin{figure}
    \centering
    \includegraphics[width=\linewidth]{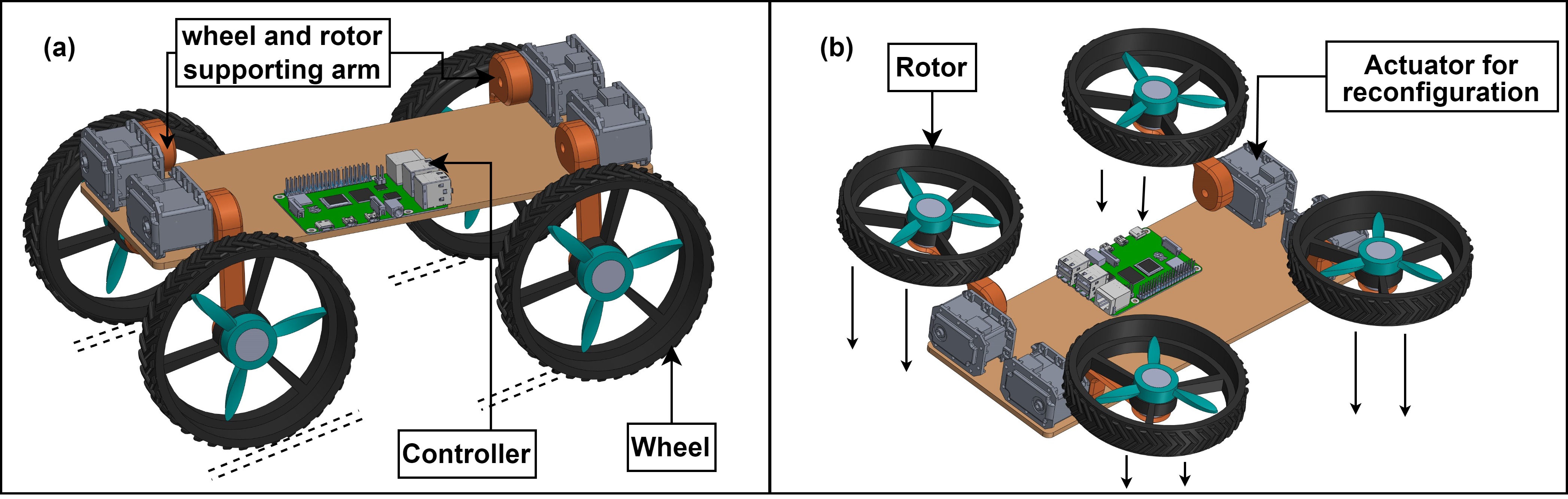}
    \caption{(a) Wheel and rotor-based multimodal robot in terrestrial configuration (b) Aerial configuration}
    \label{wheel_rotor}
\end{figure}

\begin{figure}
    \centering
    \includegraphics[width=\linewidth]{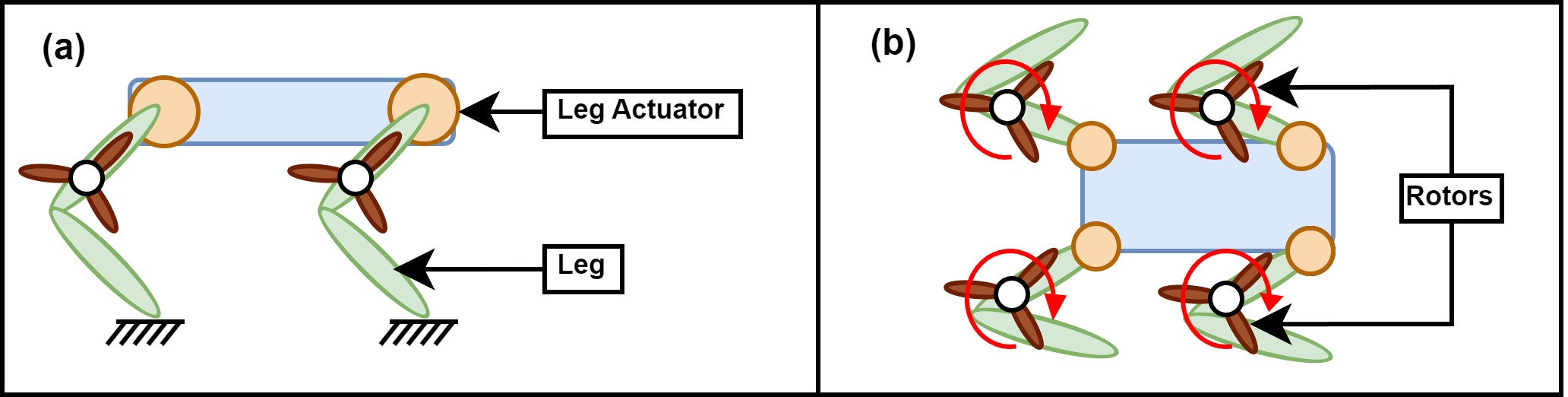}
    \caption{(a) leg with rotors in terrestrial mode (b) leg with rotors in aerial mode}
    \label{leg_rotor}
\end{figure}

\begin{figure}
    \centering
    \includegraphics[width=\linewidth]{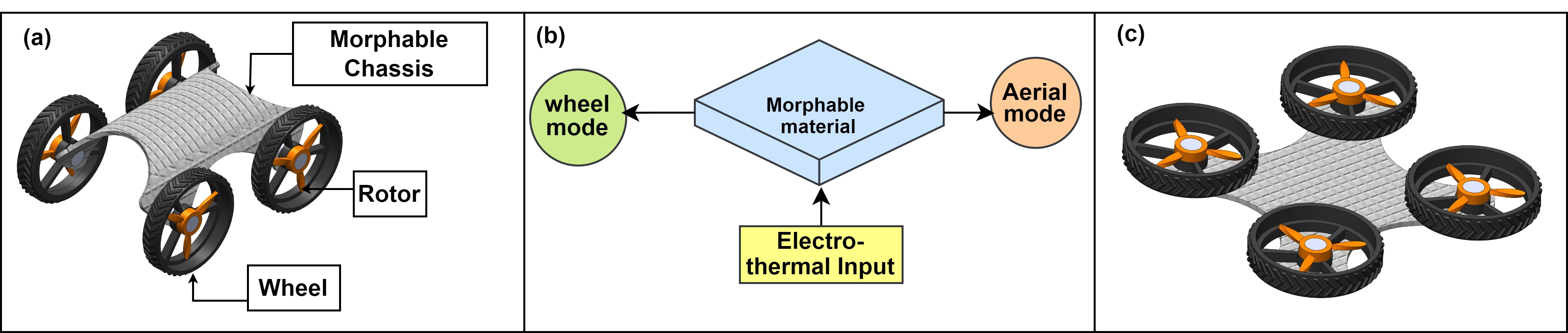}
    \caption{(a) Shape-morphable robot in terrestrial mode (b) Morph-material working (c) Aerial mode}
    \label{morphable_robot}
\end{figure}

In all the ages of mankind, mobility has always been a prime focus. It started from the wheel, which, in fact, is still on top in terms of mobility on the ground. However, there are other types of mobility requirements as well, such as in the air, on and under water, in marshy lands, in deserts, in forest lands, in snow, etc. All these different terrains and environments have their own characteristics and challenges. For such applications, the requirement is to move across a particular terrain, such as walking or rolling on the ground, flying in the air, swimming under or over the water, etc. The science of such robots is well established to a certain extent, with numerous units already deployed beyond the laboratory setting. However, there are specific applications that necessitate mobility over multiple terrains. It may involve land and air, land and water, air and water, etc. All these robots can be broadly categorized into two types based on their actuator configuration: 1. hybrid robots and 2. multimodal/reconfigurable robots. In hybrid robots, each operational mode of the system possesses its own distinct set of actuators. Consequently, in the event of actuator failure in a specific mode, alternative modes can operate safely; however, they are bulky and require more energy. In contrast, multimodal/reconfigurable robots may or may not use identical actuators across the system's operational modes (depending on the system design). Consequently, these systems offer the advantages of using fewer actuators, being compact, and being energy-efficient. However, if any actuator fails, the entire system may become inoperative. The anticipation of multimodal mobility from a robotic system poses different complexities for researchers. Like the development of an optimal system design and the selection of appropriate sensors, actuators, and control techniques for the robotic system, tailored to a specific application. Consequently, a roboticist needs to employ some unconventional design techniques and brainstorming to make such a system work. Therefore, this section outlines some robotic systems that have inspired the proposed work, alongside additional essential categories of state-of-the-art robots pertinent to the present concept. 
\paragraph{\textbf{Primary motivation:}} The proposed work is based on the ground- and aerial-based multimodal robot. Out of the two modes of operation of the system, ground locomotion is achieved through the quadruped configuration and the aerial through the intra-reconfiguration (rearrangement of the links and bodies within the system itself) of the system. Given current advancements in ground and aerial multimodal robotics, some robots, such as FCSTAR \cite{r1}, exist. This robot uses four coaxial propellers, each connected to a wheel; both propellers and wheels use the same motors. A sprawling mechanism is employed to transition between ground and airborne operations, facilitating the shift between the system's operational modes, as the authors indicate. With a similar architecture but a greater number of locomotion modes, another robot, M4 (Multi-Modal Mobility Morphobot) \cite{r3}, is reported, which is the primary motivation of the current work. This robot has walking, UGV (unmanned ground vehicle), crouching, tumbling, and mobile inverted pendulum modes on the ground, as well as a quadrotor aerial mode. Apart from this, it can also do manipulation tasks using the front or rear two articulated arms. When this robot transforms or switches from terrestrial to aerial mode, it first lands on a fixed-type landing gear, reconfigures the arm as abduction/adduction motion, and performs the aerial operation. This limitation of the robot was addressed by the ATMO robot \cite{r4}. ATMO does not land on the landing gear during the transition between two modes; rather, it reconfigures mid-air before entering the next mode. To test the system's stability in midair, the authors conducted a CFD analysis to identify the optimal rotor tilt for seamless transition between two modes. 
\paragraph{\textbf{Bio-inspired robot:}} These are such terrestrial and aerial robotic systems whose architectural detail is bio-inspired, as shown in Fig.\ref{bio_inspired_robot}. These robots can fly and crawl \cite{r5} using multiple flapping wings. Apart from using the wings solely for aerial motion, some systems have used them as fins for aquatic maneuverability \cite{r7}, along with additional appendages for terrestrial locomotion. In some cases, a single rotor is used with a lightweight stabilizer bar and a film damper \cite{r6} for stable aerial motion.
In nature, birds are the perfect example of terrestrial and aerial multimodal locomotion. To achieve bird-like flight and ground locomotion, RAVEN (Robotic Avian-inspired Vehicle for Multiple ENvironments) \cite{r16} has been developed and investigated for its potential possibilities. This robot has legs similar to birds, capable of jumping, walking, and hopping, which are required for multimodal locomotion.
\paragraph{\textbf{Wheel and rotor-based systems:}} These are the systems in which ground and aerial locomotion are achieved through the wheel and rotor configuration \cite{r13}. On the ground, wheels offer the highest energy-efficient maneuverability \cite{r8}, and in the air, rotors are definitely required. These systems can be configured by designing the wheel rim and shrouded rotors at mutually perpendicular planes \cite{r9}. The advantage of this configuration is that, while the system is in wheel mode, all rotors are idle and safe due to the external ring around them; however, in aerial mode, it is very convenient and swift due to fewer moving parts. The fundamental architecture of these robots is depicted in Fig. \ref{wheel_rotor}. Now, this wheel mode can be further extended as arc segments \cite{r10, r22}, which, when in the same phase, constitute a full wheel and, when out of phase, allow the aerial mode. Another configuration of the segmented wheel can also work as a legged robot \cite{r14}. In some systems, instead of mounting these rotors on wheels, they can be mounted on the main body of the system \cite{r11} and do not require intra-reconfiguration during switching from the terrestrial mode to the aerial mode. On the other hand, these rotors can be deployed via intra-reconfiguration of the system \cite{r12}. In multimodal robots, interventions or manipulations on the ground or in the air are also important. In aerial mode, the primary concern is system stability. Furthermore, incorporating midair intervention may complicate the operation. Nonetheless, this can be accomplished with the appropriate system design \cite{r15} and an appropriate control scheme.
\paragraph{\textbf{Legs with rotors:}} Using legged robots along with rotors is really a complex task to achieve. Since attaching legs and rotors (or thrusters, i.e., ducted fans) is a complex process from both mechanical and control perspectives. To achieve this, some robots use these rotors near the hip joint \cite{r17} or near the knee joint \cite{r20}. Since attaching these rotors away from the center of gravity of the overall system may pose a challenge in balancing the system in both modes, a basic arrangement of this combination is depicted in Fig. \ref{leg_rotor}. In another variation, a quadruped-based legged robot is equipped with the four thrusters (ducted fans) \cite{r18} required for aerial locomotion. However, the selection of thrusters and rotors for quadrupeds is entirely dependent on thrust requirements and the mechanical complexity of executing real-world motion. Hence, in some cases, these rotors are directly mounted on top of the quadruped body itself \cite{r19}, to accommodate other hardware components required for the operation. Most robots in this category feature a full quadruped body, augmented by four rotors or thrusters of varying sizes and positions. All the joints of the quadruped configuration have their own joint motors, and the aerial has its own rotors or thrusters. But there are very few systems in which both legged and aerial configurations are fully achieved with the same number of actuators, such as SPIDAR \cite{r21}. In this system, a quadruped configuration is fully achieved by small thrusters that are also used for aerial motion.
\paragraph{\textbf{Smart morphable material-based robot:}} These are a very special class of reconfigurable, multimodal robots. Unlike conventional methods, which use rigid bodies and actuators at the joints, robots switch between modes due to relative motion between the links. These types of robots use a shape-morphable material underneath. These materials are electro-thermally excited \cite{r23, r24} to yield a particular shape or geometry. The basic working of these types of multimodal robots can be understood from the Fig. \ref{morphable_robot}. Now, since a particular desired shape of the material is a function of electrical input, it can therefore be programmed for a definite shape. In general, these smart materials have been used to make the system's chassis. So, when the system needs to operate in terrestrial mode, the underlying chassis is rigid, and its rigidity is controlled by a programmed electro-thermal input. Now, if the system requires a change in mode of operation, a different input is provided to the chassis material, and it changes its shape to the required configuration for that mode.
\paragraph{\textbf{Current landscape and objectives of the proposed work:}}From the state of the art, it is observed that the ground and aerial-based multimodal robot is investigated mostly in the direction of wheel- and rotor-based systems. Moreover, in the field of legged and aerial systems, there is limited work. Therefore, the authors decided to contribute to the development of legged and rotor-based multimodal robots by implementing a step-by-step modeling approach and testing the system on hardware at every stage. With this central idea, this work is primarily based on the system modeling (kinematic modeling only) of the locomotor required for the full multimodal quadruped system for terrestrial and aerial operation. The current work is the first part of the full system, in which the overall system at the kinematic level is described, followed by workspace analysis, singularity analysis, and gait generation, with the help of a single-leg-based test rig setup. So with this fact, the present work can be materialized with the following key objectives, which are as follows:
\begin{enumerate}
    \item Forward and inverse kinematics of the overall robot, using a single leg connected to the floating base.
    \item Kinematic analysis of the double closed-loop-based single leg.
    \item workspace and singularity analysis of the leg.
    \item Development of the gaits for the full system using a single leg only.
    \item Implementation of the different planned gaits on the real hardware using a single leg test rig for the locomotor module.
\end{enumerate}
The paper is divided into different sections. The main proposed system is explained in Section 2. The development of kinematic modeling using non-standard DH parameters is explained in Section 3. Formulation of the leg Jacobian is discussed in Section 4. Workspace generation and singularity analysis are discussed in Section 5. Hardware realization and its connection to the system's modeling and simulation are discussed in Section 6. Finally, the introduction of the next phase of the work and the future scope of the present work are explained in Section 7.
\section{Description of the proposed system}
\begin{figure}
    \centering
    \includegraphics[width=0.85\linewidth]{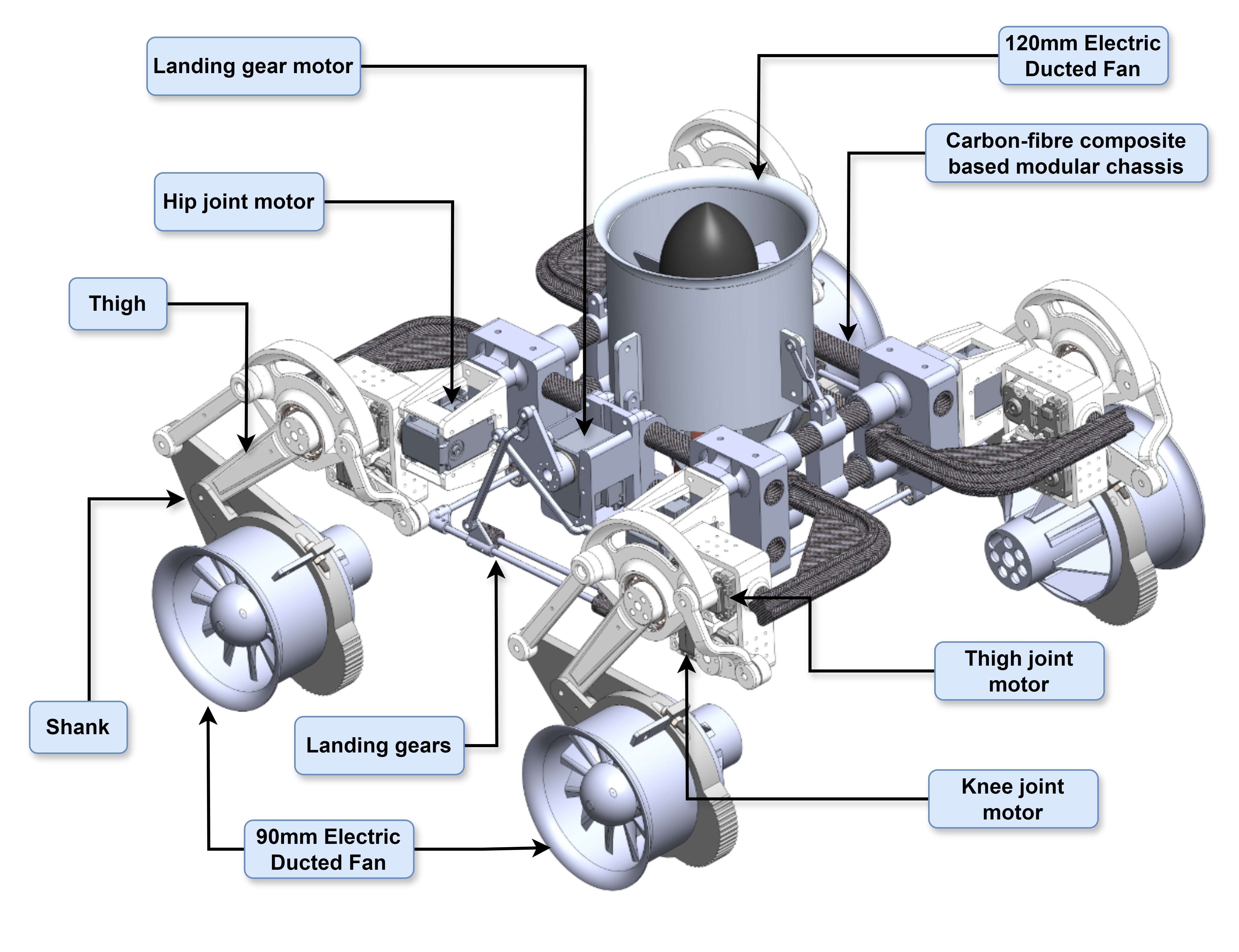}
    \caption{CAD model of the proposed design of the multimodal reconfigurable robot}
    \label{robot_description}
\end{figure}

In the proposed system, the main objective is to explore the dynamics of a reconfigurable robot that can transform from a terrestrial mode (walking, quadruped) to an aerial mode (flying, multi-rotor). As a starting point for the study, it was very important to use a specific application as a use case, which can provide definite information about the system's shape, size, and functional features. This information is required for the mathematical modeling and analysis of the system. Therefore, certain applications were considered to conceptualize the overall system. These applications include post-disaster response, medical supplies for civil and armed forces personnel during war or similar critical situations, and tactical missions of the armed forces, among others.\\ Therefore, in the purview of these applications, a small-scale, desktop-sized robotic system is envisioned as a starting point for the proposed work, as shown in Fig. \ref{robot_description}.
From the same figure, it is clear that the proposed robot has four legs, which, in the current context, are referred to as locomotor legs. These locomotors are responsible for the robot's ground and aerial motion. All four locomotors are identical in shape, size, and configuration. Each one of them is equipped with 3 joints: hip abduction/adduction, thigh flexion/extension, and knee flexion/extension. For each of these joints, there is a geared servo motor to actuate it. In addition to these joint motors, a 90 mm Electric Ducted Fan (EDF) is connected to the shank link. These EDFs are required only for aerial motion; during ground locomotion, they will be idle and do not contribute to the quadruped mode, except as additional weight. On the floating base, there is a 120 mm EDF connected for the aerial locomotion to assist the other four leg-mounted spatial EDFs in order to compensate for the thrust required during the takeoff and landing. A detailed explanation of the functioning of each locomotor, along with its synchronization with the centrally mounted EDF, is provided in the next sub-section.
\subsection{Operation sequence of the proposed robot}
\begin{figure}
    \centering
    \includegraphics[width=\linewidth]{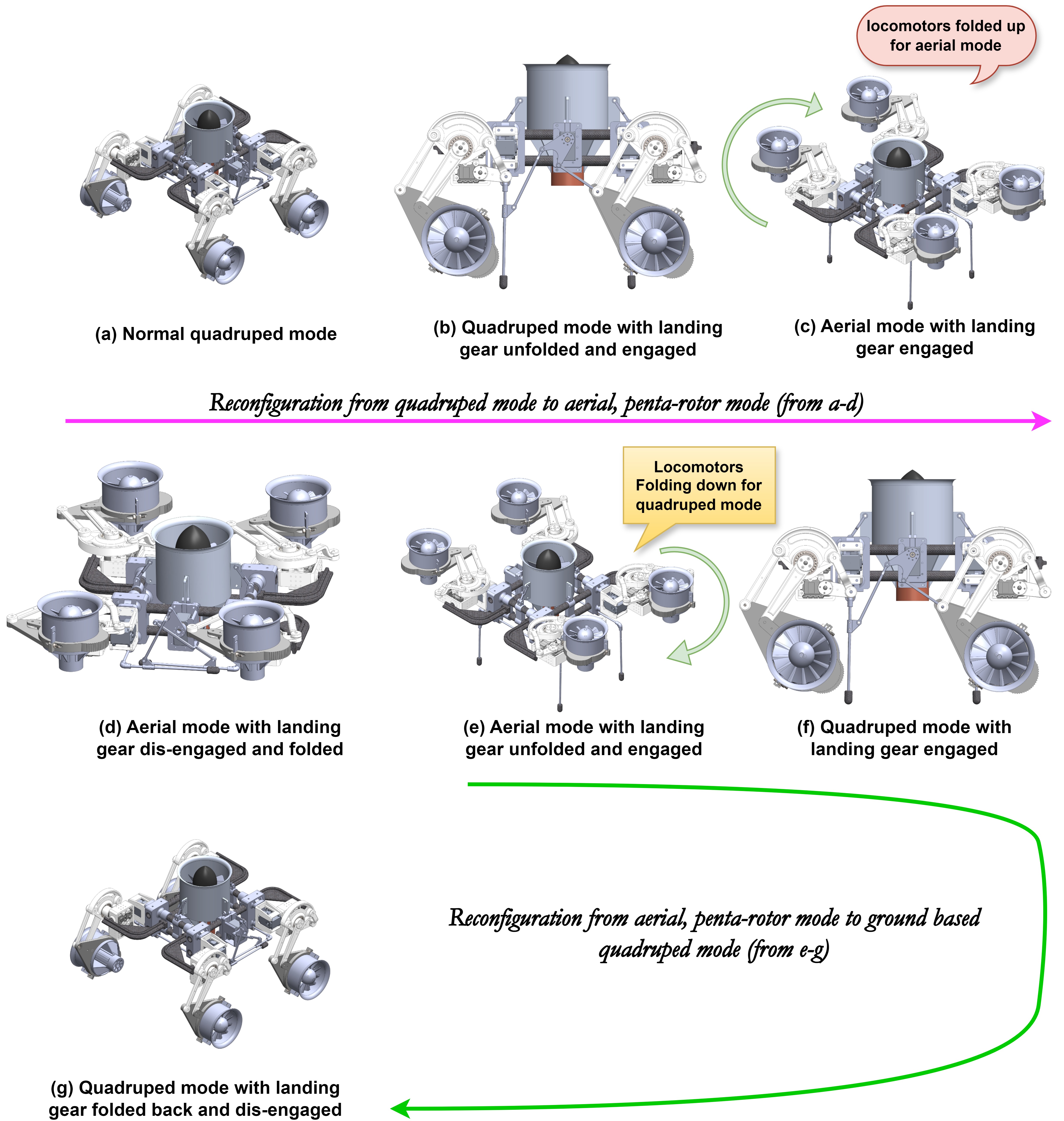}
    \caption{Transformation stages of reconfiguration between quadruped mode and pentarotor aerial mode}
    \label{transformation_stages}
\end{figure}

Since the proposed system is multimodal and includes articulation, it can switch between modes. Switching between the modes is accomplished via multiple intermediate stages, shown in Fig.~\ref{transformation_stages}. This switching eventually enables the system to ascend from the ground to the air or descend from the air to the ground. When the system is on the ground, it is in quadruped mode; when in the air, it is in pentarotor aerial mode. The switching of both modes is explained below in terms of the takeoff and landing sequence.
\subsubsection{Take off sequence} From the Fig. \ref{transformation_stages}, it can be observed that, in stage (a), the robot is in the state of quadruped mode for ground operations. To switch from quadruped-terrestrial mode to pentarotor-aerial mode, the landing gear must first be engaged, as shown in Fig. \ref{transformation_stages}b. The purpose of this landing gear is to facilitate the system's transition between two modes. Therefore, when the landing gear is engaged, the whole system (except the landing gear module itself) will be at rest/idle and supported on these gears. Now, all four locomotors can move in an upward direction as shown in the Fig. \ref{transformation_stages}c. Once these locomotors are completely folded and supported on the landing gears, all four rotors (EDFs), including the main central EDF, can be powered up to produce the required upward thrust for takeoff. After takeoff from the ground, landing gears are not required; hence, they will be folded up, completing the takeoff sequence, as shown in Fig. \ref{transformation_stages}d. The centrally mounted EDF compensates for thrust requirements during takeoff and when any of the four locomotor-mounted EDFs fail.
\subsubsection{Landing sequence}  When the system is in aerial mode and required to switch into ground-based quadruped mode, again, landing gears will be used to facilitate this switching. First, the landing gear will be deployed to prepare the system for landing. Also, during this stage, the system will be hovering near the ground at a safe height to avoid any collision with the ground or with any other object near the ground, as shown in Fig. \ref{transformation_stages}e. These fully extended landing gears help the system to safely land on the ground. During the onset of landing, all EDFs' rpm will be slowly decreased with a corresponding fall in thrust magnitude. When the landing gear finally touches the ground safely, all the EDFs will be turned off. Now, all four locomotors will move in a downward direction to assume the quadruped pose, as shown in Fig. \ref{transformation_stages}f. Since the system is back in quadruped mode and ready for ground locomotion, landing gears are no longer required; they will be folded up to complete the landing sequence.
\section{Kinematic modeling of the system} In the previous section, different modes and the functioning of the system are described. However, to build the physical system, it is necessary to understand its mathematical model, which, in turn, helps understand the overall system's dynamics. The mathematical model of the system typically requires kinematic analysis, followed by its dynamic analysis. Therefore, in this section, the system's kinematics is first described, laying the foundation for the dynamic analysis, which will be carried out in the next work, the continuation of the present work.
\subsection{Forward kinematics} 
\begin{figure}
    \centering
    \includegraphics[width=\linewidth]{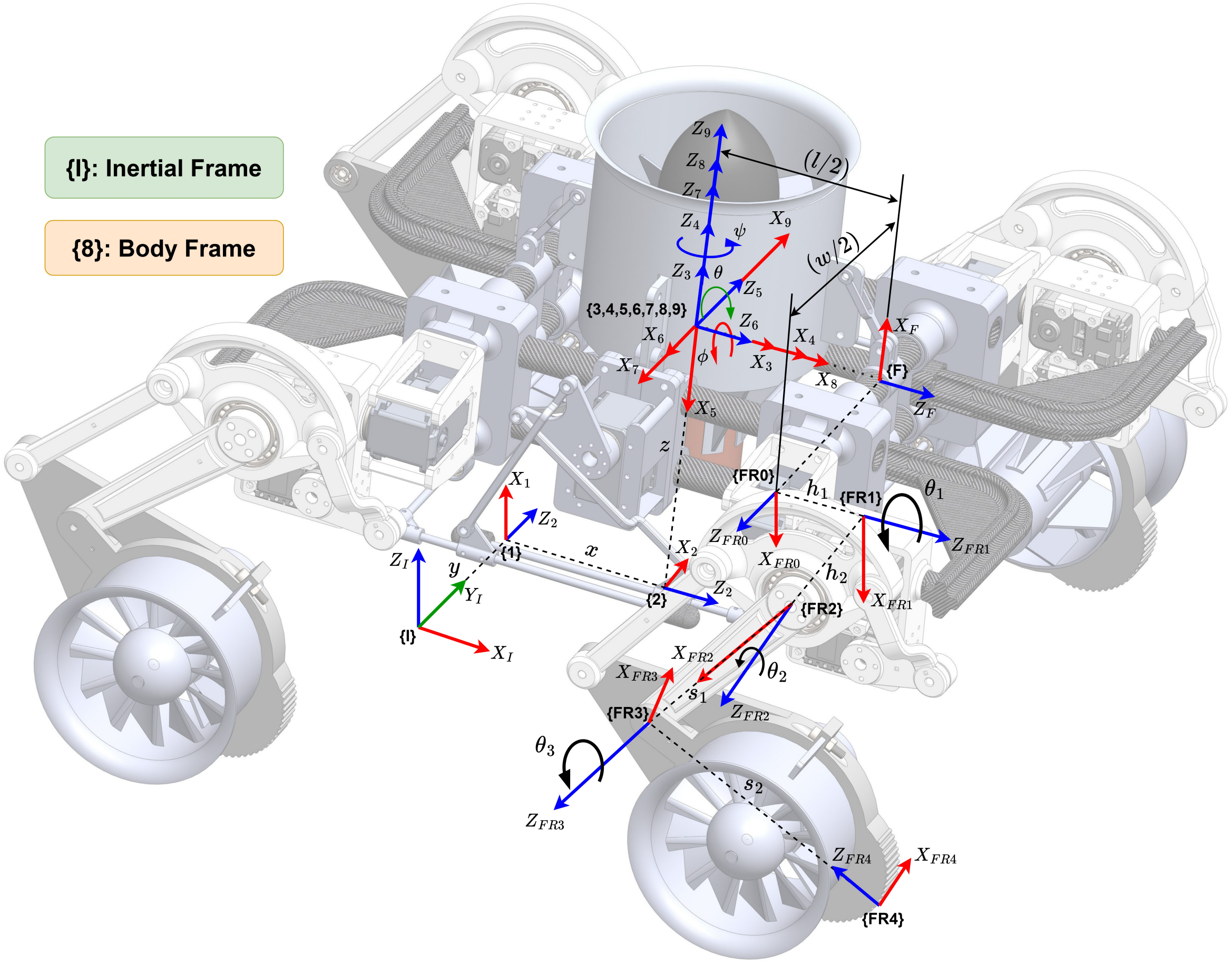}
    \caption{DH Table of quadruped for front right locomotor}
    \label{DHT_quadruped}
\end{figure}

The proposed system, as described in the introduction section, is reconfigurable in nature. Therefore, to understand the motion of the system in different modes, its kinematic analysis is carried out with the help of a non-standard DH parameter table, as depicted in Table \ref{table_1}, with the corresponding kinematic diagram in Fig. \ref{DHT_quadruped}.
\begin{table}[htbp]
\centering
\caption{DH parameter table of the quadruped for the front-right locomotor.}
\label{table_1}
\renewcommand{\arraystretch}{1.2}
\begin{tabular}{ccccc}
\hline
Frame ID & $\alpha_{i-1}$ (rad) & $a_{i-1}$ (m)
& $\theta_i$ (rad) & $d_i$ (m) \\
\hline
1 & $3\pi/2$ & $0$ & $3\pi/2$ & $y$ \\
2 & $3\pi/2$ & $0$ & $3\pi/2$ & $x$ \\
3 & $3\pi/2$ & $0$ & $3\pi/2$ & $z$ \\
4 & $0$ & $0$ & $\psi$ & $0$ \\
5 & $3\pi/2$ & $0$ & $\pi/2+\theta$ & $0$ \\
6 & $\pi/2$ & $0$ & $3\pi/2+\phi$ & $0$ \\
7 & $\pi/2$ & $0$ & $0$ & $0$ \\
8 & $0$ & $0$ & $\pi/2$ & $0$ \\
9 & $0$ & $0$ & $\pi/2$ & $0$ \\
F & $3\pi/2$ & $0$ & $\pi/2+\theta$ & $l/2$ \\
FR$_0$ & $3\pi/2$ & $0$ & $\pi$ & $w/2$ \\
FR$_1$ & $3\pi/2$ & $0$ & $\theta_1$ & $h_1$ \\
FR$_2$ & $\pi/2$ & $0$ & $\theta_2$ & $h_2$ \\
FR$_3$ & $0$ & $s_1$ & $\pi/2+\theta_3$ & $0$ \\
FR$_4$ & $3\pi/2$ & $0$ & $0$ & $-s_2$ \\
\hline
\end{tabular}
\end{table}
Only one DH table for the front right locomotor is shown here; however, analogous DH tables are constructed for the other locomotors, which are omitted for conciseness. It can be seen from Table \ref{table_1} that the system has 12 controllable inputs, corresponding to 3 active joints per locomotor, which, in turn, determine the 6 degrees of freedom of the floating base. These degrees of freedom are in terms of 3 translation motions, as represented by $x$, $y$, and $z$ in the table, and 3 rotational motions $\psi$, $\theta$, and $\phi$ about $X$, $Y$, and $Z$ axes with respect to the inertial frame \{I\}. There are other geometrical parameters of the system, which are required for the kinematics of the system; these are provided in the Table \ref{table_2}.
\begin{table}[htbp]
\centering
\caption{Geometrical parameters of the system.}
\label{table_2}

\renewcommand{\arraystretch}{1.2}

\begin{tabular}{c c l}
\hline
Parameter & Magnitude & Description \\
\hline
$l$     & 0.19 m & Length of the robot \\
$w$     & 0.14 m & Width of the robot \\
$h_1$   & 0.04 m & Length between frames $\{FR_0\}$ and $\{FR_1\}$ \\
$h_2$   & 0.05 m & Length between frames $\{FR_1\}$ and $\{FR_2\}$ \\
$s_1$   & 0.11 m & Length of the thigh link \\
$s_2$   & 0.14 m & Length of the shank link \\
\hline
\end{tabular}
\end{table}
Now, in order to find the foot position, which is represented by the frame \{FR4\} in Fig. \ref{DHT_quadruped} and Table \ref{table_1}, forward kinematics can be used.
Which is the position of the foot for a given set of joint angles of the locomotor and can be evaluated by the homogeneous transformation of Frame \{FR4\} with respect to the frame \{I\}, as represented in equation \eqref{eq_1}. 
\begin{equation}
  \begin{bmatrix}
      P_{x}\\P_{y}\\P_{z}
  \end{bmatrix}=\begin{bmatrix}
      h_{1} + x + s_{1}\cos{\theta_{2}}\sin{\theta_{1}} + s_{2}\cos{\theta_{1}}\sin_{3} + s_{2}\cos{\theta_{2}}\cos{\theta_{3}}\sin{\theta_{1}}\\
      y - h_{2} + s_{1}\sin{\theta_{2}} + s_{2}\cos{\theta_{3}}\sin{\theta_{2}}\\
      z - s_{1}\cos{\theta_{1}}\cos{\theta_{2}} + s_{2}\sin{\theta_{1}}\sin{\theta_{3}} - s_{2}\cos{\theta_{1}}\cos{\theta_{2}}\cos{\theta_{3}}
  \end{bmatrix}
  \label{eq_1}
\end{equation}
Where, $[P_{x},P_{y},P_{z}]\textsuperscript{T}$ is the position vector of the foot with respect to the inertial frame, provided the yaw($\psi$), pitch($\theta$) and roll($\phi$) angles are set to zero, to make the expression compact, however, the full expression is also provided in appendix \ref{forward_kinematics}. This section establishes the relationship between the robot's joint and task spaces via a homogeneous transformation. The robot's foot position can be determined based on specified joint angles. Which is simple and direct from a computational perspective. To enable the robot's locomotion to adhere to a specified trajectory, it is necessary to correlate the task space with the joint space. In other words, identifying the joint configuration for a given task space (discrete trajectory points). This mapping can be achieved through inverse kinematics, as discussed in the next section.
\subsection{Inverse kinematics}
\begin{figure}
    \centering
    \includegraphics[width=0.5\textwidth]{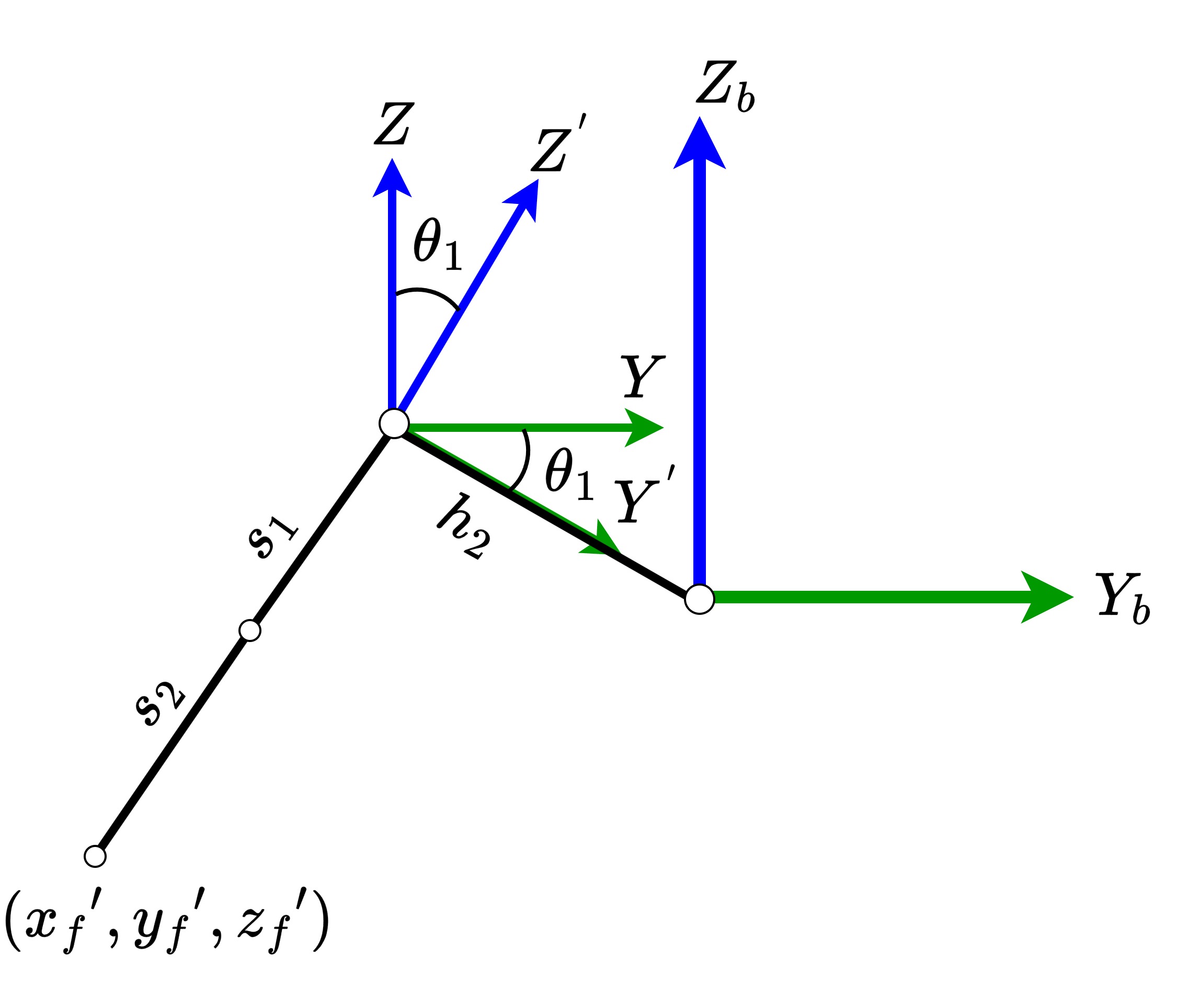}
    \caption{Effect of the joint angle $\theta_{1}$ on the pose of the leg in front view}
    \label{IK_2}
\end{figure}

In this section, for a given leg position in task space, the required joint-space angles will be evaluated using leg geometry. Calculation of the joint angles can also be done by solving equation \eqref{eq_1}, but due to the nonlinear nature of this equation, it is difficult to obtain a closed-loop solution. However, there are still numerical techniques for solving these equations for the joint angles, but computationally, it will be costly. Therefore, to obtain a closed-loop expression, the geometrical method is used and discussed in this section.
\begin{figure}
    \centering
    \includegraphics[width=1.1\textwidth]{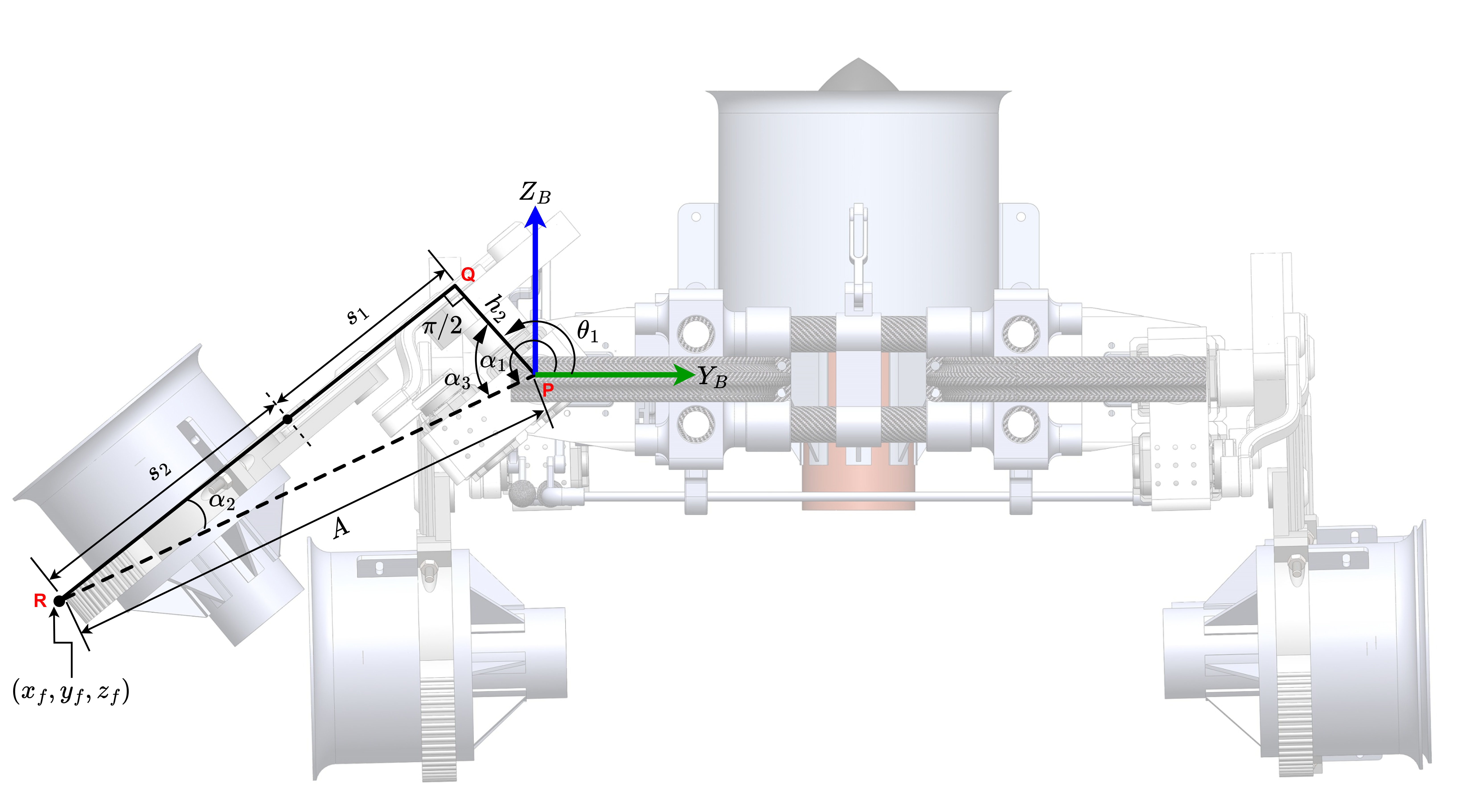}
    \caption{Front view of the leg to calculate first joint angle, $\theta_{1}$}
    \label{IK_1}
\end{figure}
In Fig. \ref{IK_1}, a side view of the front right leg of the robot is shown, in which $\theta_{1}$ is the hip adduction/abduction angle. Additionally, the reference frame chosen for this calculation is the robot's body frame itself and is shown at the hip location for calculation purposes; however, it is located as shown by frame \{8\} in Fig. \ref{DHT_quadruped}. Once this angle is calculated, it will be recalculated as per the DH relation between the body frame and the corresponding hip frame.\\
From Fig. \ref{IK_1}, in $\triangle$PQR:
\begin{equation}
    A = \sqrt{y_{f}^{2}+z_{f}^{2}}
    \label{eq_2}
\end{equation}
where $x_{f}, y_{f}$ and $z_{f}$ are the coordinates of the foot with respect to the body frame.
\begin{equation}
\alpha_{1} = \arctan2(z_{f},\, y_{f})
\label{eq_3}
\end{equation}
\begin{equation}
    \frac{\sin{\alpha_{2}}}{h_{2}} = \frac{\sin({\pi/2})}{A}; \hspace{1cm} \alpha_{2} = \arcsin\!\left( \frac{h_{2}\sin(\pi/2)}{A} \right)
    \label{eq_4}
\end{equation}
\begin{equation}
    \alpha_{3}=\pi/2-\alpha_{2}
    \label{eq_5}
\end{equation}
\begin{equation}
    \theta_{1}=(\alpha_{1} \pm \alpha_{3})
    \label{eq_6}
\end{equation}
In equation \eqref{eq_6}, the (+) sign is for the front and rear legs of the left side of the robot, and the (-) sign is for the front and rear legs of the right side of the robot. As in the current case, the front right leg is involved, so the (-) sign will be used. After calculating $\theta_{1}$, Fig. \ref{IK_1} can be referred to for calculating angles $\theta_{2}$ and $\theta_{3}$. In the figure, it can be seen that, due to the angle $\theta_{1}$, the leg has changed pose, and the foot's coordinates have also changed. Therefore, to accommodate the angle $\theta_{1}$ in the calculation of the other two angles $\theta_{2}$ and $\theta_{3}$, the reference frame needs to be rotated about the x-axis by an angle of $\theta_{1}$ in opposite to the actual rotation, as shown in Fig. \ref{IK_2}, due to which the new coordinates of the foot $[x_{f}{'},y_{f}{'},z_{f}{'}]\textsuperscript{T}$ can be calculated as follows:
\begin{equation}
    \begin{bmatrix}
        x_{f}{'}\\ y_{f}{'}\\ z_{f}{'}
    \end{bmatrix}=Rot_{x}(-\theta_{1})\begin{bmatrix}
        x_{f}\\y_{f}\\z_{f}
    \end{bmatrix}
    \label{eq_7}
\end{equation}
Since new transformed foot coordinates are obtained from \eqref{eq_7}. Using these coordinates, the other two angles can be calculated as follows.\\
From Fig. \ref{IK_3}, in $\triangle$DEF:
\begin{figure}
    \centering
    \includegraphics[width=0.75\textwidth]{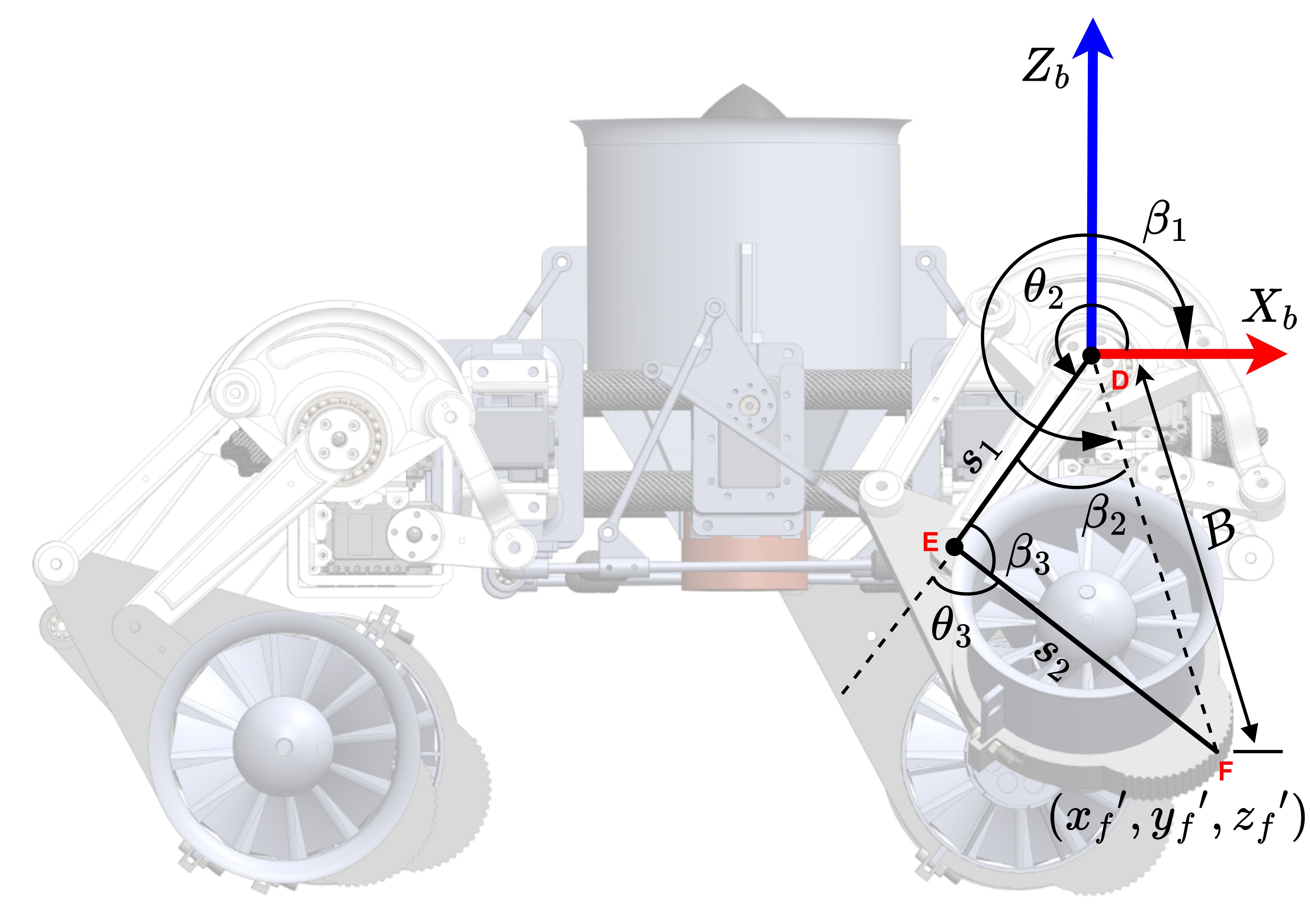}
    \caption{Side view of the leg to calculate the angles, $\theta_{2}$ and $\theta_{3}$}
    \label{IK_3}
\end{figure}
\begin{equation}
     DF^2= \sqrt{DE^2+EF^2}
     \label{eq_8}
\end{equation}
\begin{equation}
    B = \sqrt{x_{f}{'}^{2}+z_{f}{'}^{2}}
    \label{eq_9}
\end{equation}
\begin{equation}
    \beta_{1}=\arctan2\!\left(\frac{z_{f}{'}}{x_{f}{'}}\right);\hspace{1cm} 0\leq \beta_{1} \leq 2\pi
    \label{eq_10}
\end{equation}
\begin{equation}
   \beta_{2} = \arccos\left( \frac{s_{1}^{2} + B^{2} - s_{2}^{2}}{2 s_{1} B} \right)
    \label{eq_11}
\end{equation}
\begin{equation}
    \theta_{2} = (\beta_{1} - \beta_{2})
    \label{eq_12} 
\end{equation}
\begin{equation}
  \beta_{3} = \arccos\left( \frac{s_{1}^{2} + s_{2}^{2} - B^{2}}{2 s_{1} s_{2}} \right)
  \label{eq_13}
\end{equation}
\begin{equation}
    \theta_{3}=(\pi-\beta_{3})
    \label{eq_14}
\end{equation}
From equations \eqref{eq_6}, \eqref{eq_12}, and \eqref{eq_14}, all three joint angles are calculated corresponding to a given or desired foot position. These joint angles will later be used to actuate the joint-based trajectory for each leg of the robot. Additionally, in this section, the inverse kinematics of the robot's front right leg is solved, and the same approach will be used for the rear right leg as well. However, for the front left leg and rear left leg, the value of the angle $\theta_{1}$ will be different, as discussed in equation \eqref{eq_6}, but for the other two angles $\theta_{2}$ and $\theta_{3}$, it will be the same. 
\subsection{Adding the effect of floating base rotation in the inverse kinematics}
In the previous section, for a given foot position of the robot, its joint angles were calculated using the geometrical method. Using these joint angles, the actuators of each leg receive the actuation command and ultimately dictate the robot's overall pose. But in some situations, the robot's floating base also needs to assume a specific pose using roll ($\phi$), pitch ($\theta$), and yaw ($\psi$), without affecting any of the legs' positions. In this section, the leg's foot coordinates will be calculated based on the required roll, pitch, and yaw angles, along with the center of rotation of the floating base. After calculating these foot coordinates, inverse kinematics, as described in the previous section, can be applied directly.
\begin{figure}
    \centering
    \includegraphics[width=0.7\textwidth]{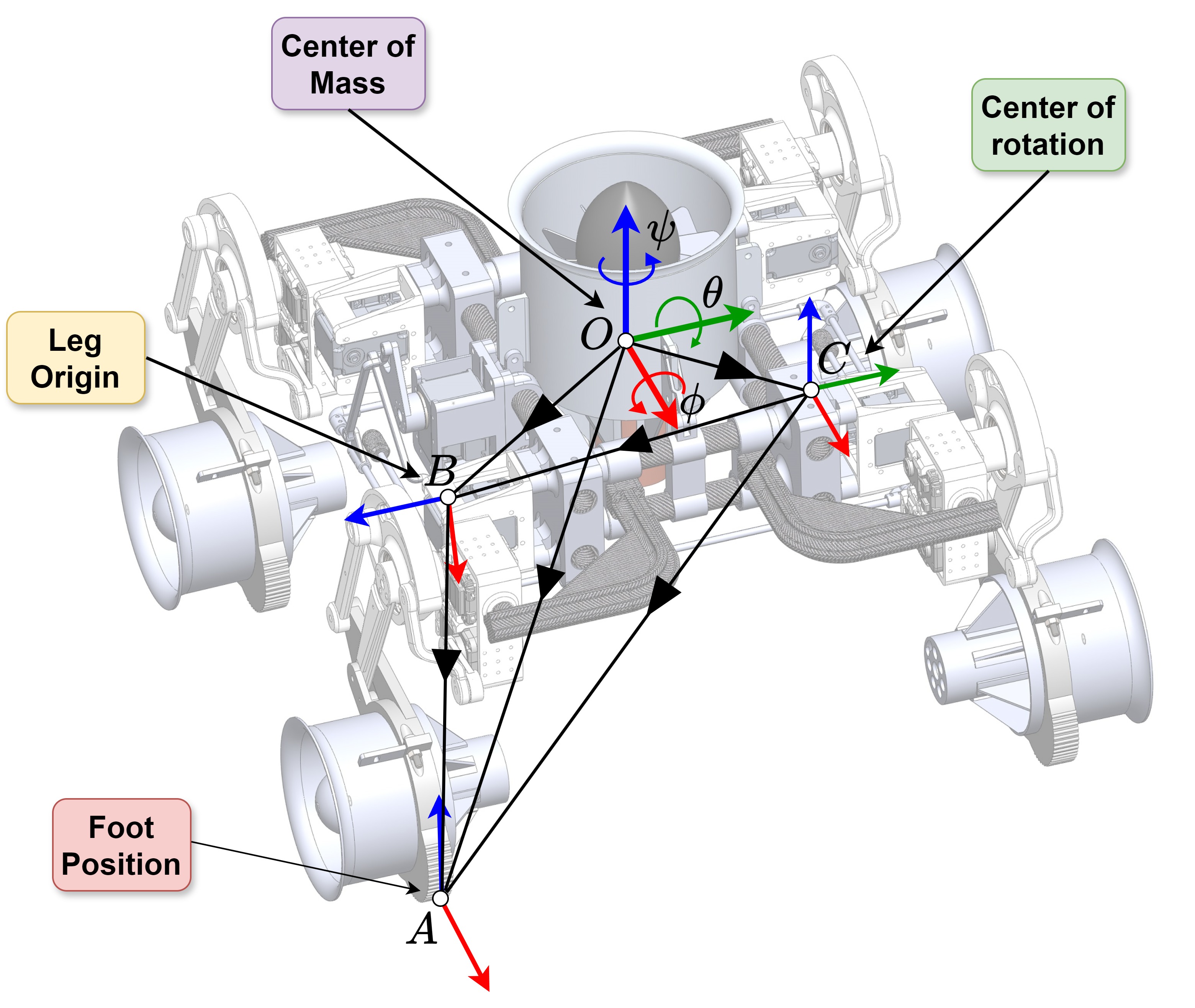}
    \caption{Effect of rotation of the floating base}
    \label{rpy}
\end{figure}
From Fig. \ref{rpy}, the vector relation between the foot position (A), leg origin (B), center of rotation (C), and center of mass of the robot (O) is given by:
\begin{equation}
    \Vec{CA}=\Vec{CB} + \Vec{BA}
    \label{eq_15}
\end{equation}
where,
\begin{equation}
    \Vec{CB} = \Vec{OB} - \Vec{OC}
    \label{eq_16}
\end{equation}
\eqref{eq_15} gives the position of the foot with respect to the center of rotation, without including floating base rotation. If the center of mass and the center of rotation coincide, then the rotation is with respect to the body frame of the system. Additionally, point '\textbf{O}' also represents the body frame of the system, which includes all three: roll, pitch, and yaw angles. Therefore, to include these rotations in equation \eqref{eq_16}, it needs to be pre-multiplied by a rotation matrix \text{R}\textsuperscript{-1}, which is given by equation \eqref{eq_17}.
\begin{equation}
    \text{R} = \begin{bmatrix}
        \cos{\psi}\cos{\theta} & -\sin{\psi}\cos{\phi} + \cos{\psi}\sin{\theta}\sin{\phi} & \sin{\psi}\sin{\phi}+\cos{\psi}\cos{\phi}\sin{\theta}\\
        \sin{\psi}\cos{\theta} & \cos{\psi}\cos{\phi}+\sin{\phi}\sin{\theta}\sin{\psi} & -\cos{\psi}\sin{\phi}+\sin{\theta}\sin{\psi}\cos{\phi}\\
        -\sin{\theta} & \cos{\theta}\sin{\phi} & \cos{\theta}\cos{\phi}
        \label{eq_17}
    \end{bmatrix}
\end{equation}
Where R is the transformation function, which transforms the coordinates from the body frame to the inertial frame. Now using \eqref{eq_17} in equation \eqref{eq_15}:
\begin{equation}
    \Vec{CA}=\text{R}\textsuperscript{-1}\left(\Vec{CB} + \Vec{BA}\right)
    \label{eq_18}
\end{equation}
Equation \eqref{eq_18} is the complete equation to find the foot position with respect to the center of rotation of the robot, including floating base rotation. Additionally, the foot coordinates with respect to the robot's body frame are calculated in \eqref{eq_18}, which are the required coordinates for the inverse kinematics, as discussed in the previous section. 
\begin{equation}
    \Vec{BA}=\Vec{CA}-\Vec{CB}
    \label{eq_19}
\end{equation}
\begin{figure}
    \centering
    \includegraphics[width=0.65\linewidth]{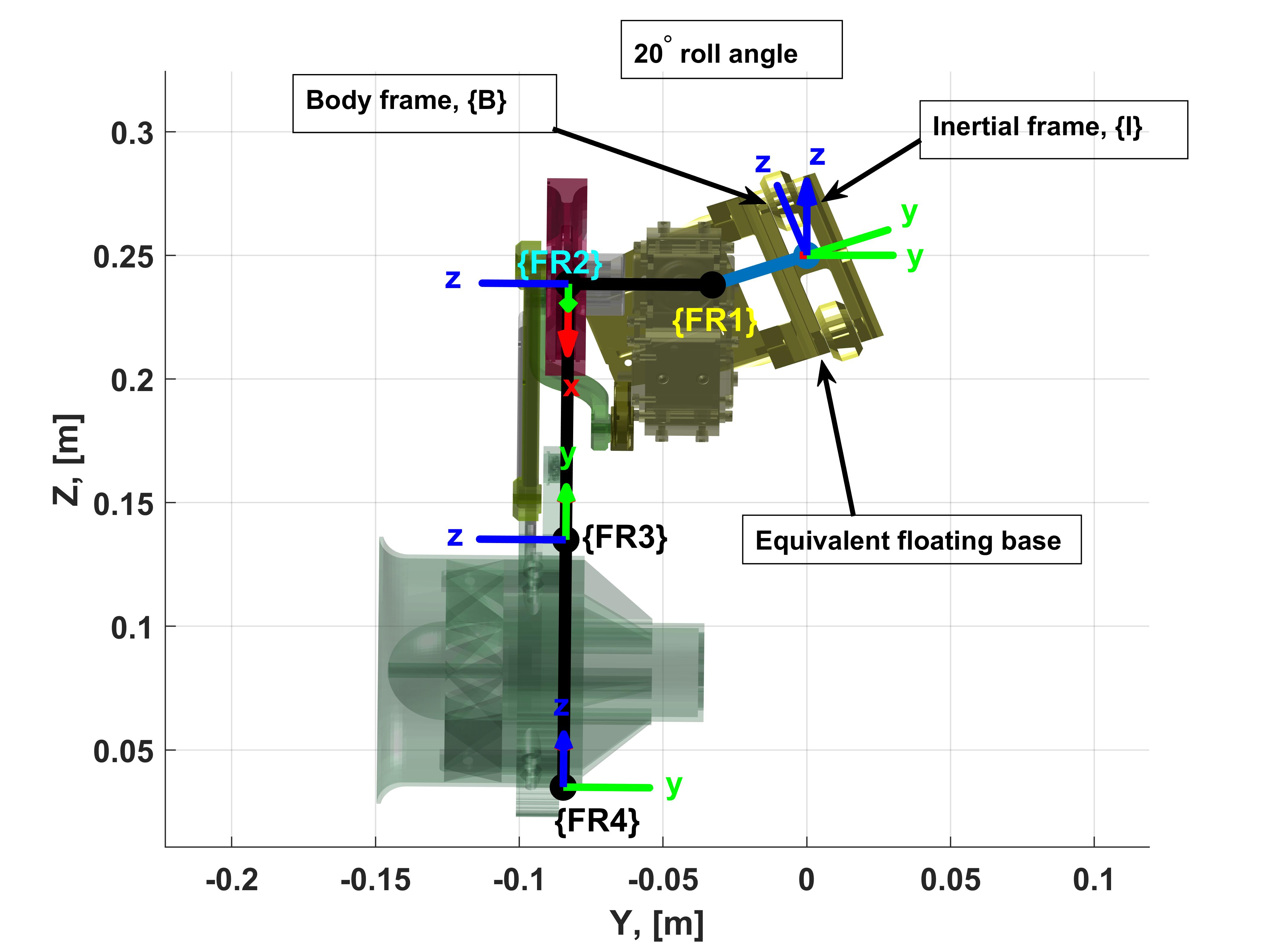}
    \caption{Leg's pose with roll offset implementation}
    \label{roll_offset}
\end{figure}
\begin{figure}
    \centering
    \includegraphics[width=0.65\linewidth]{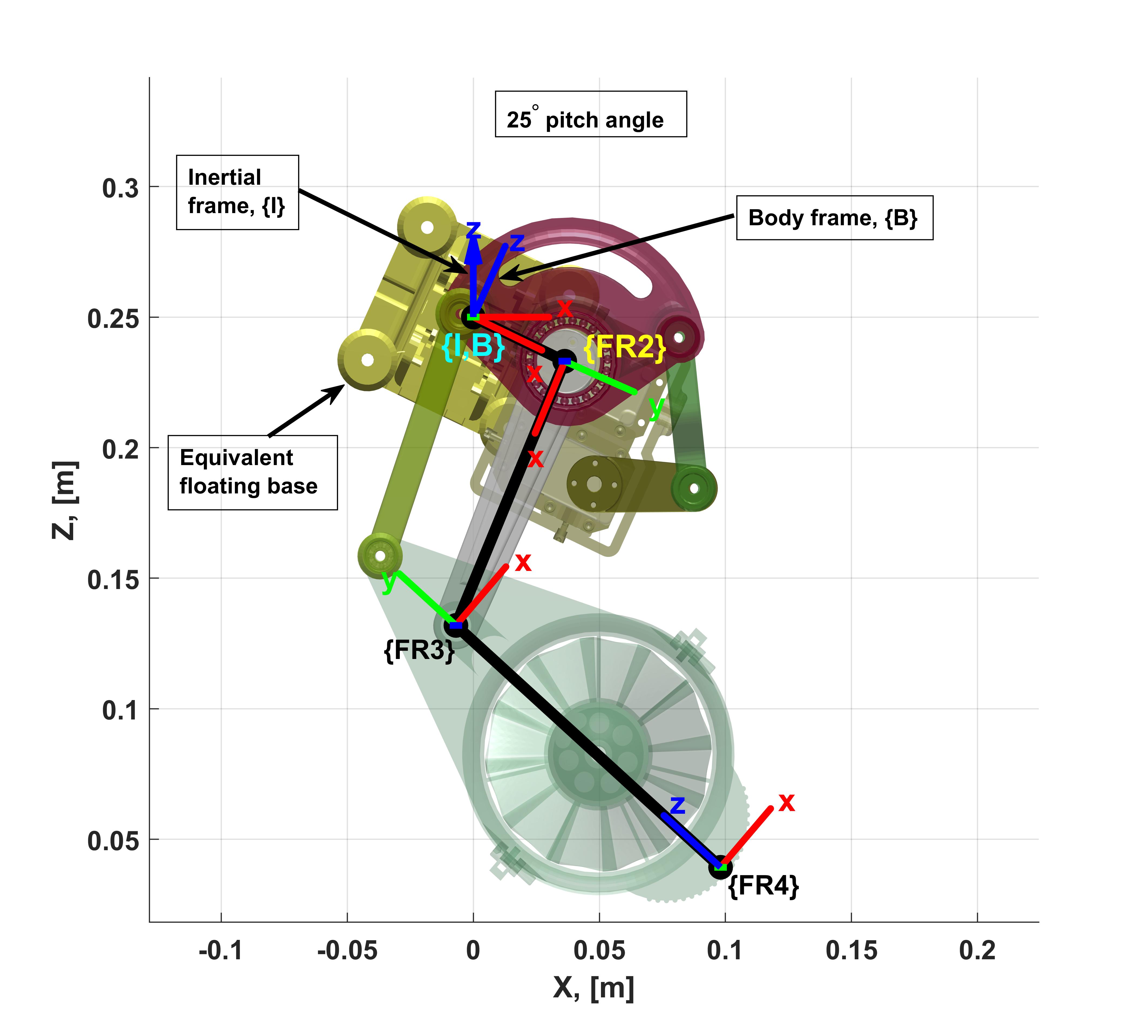}
    \caption{Leg's pose with pitch offset}
    \label{pitch_offset}
\end{figure}
To illustrate the effect of floating base rotation on the robot's body, it is explained here with the help of a single leg, as shown in Fig. \ref{roll_offset} and Fig. \ref{pitch_offset}. For this illustration, the center of rotation is taken to be $\{0, 0, 0\}$, meaning the center of rotation is the body frame itself, represented by point '\textbf{O}' in Fig. \ref{rpy}. From both these figures, it can be observed that, due to the roll or pitch angle provided to the system, only the robot's floating base rotates by that angle. However, the leg's joint will be adjusted accordingly for these rotations in order to keep the foot frame $\{\textbf{FR4}\}$ fixed on the ground. These offsets help shift the robot's center of mass as it ascends or descends over undulating terrain, maintaining system stability.
\subsection{Simplification of the system model using a single-leg test rig}
\begin{figure}
    \centering
    \includegraphics[width=0.7\textwidth]{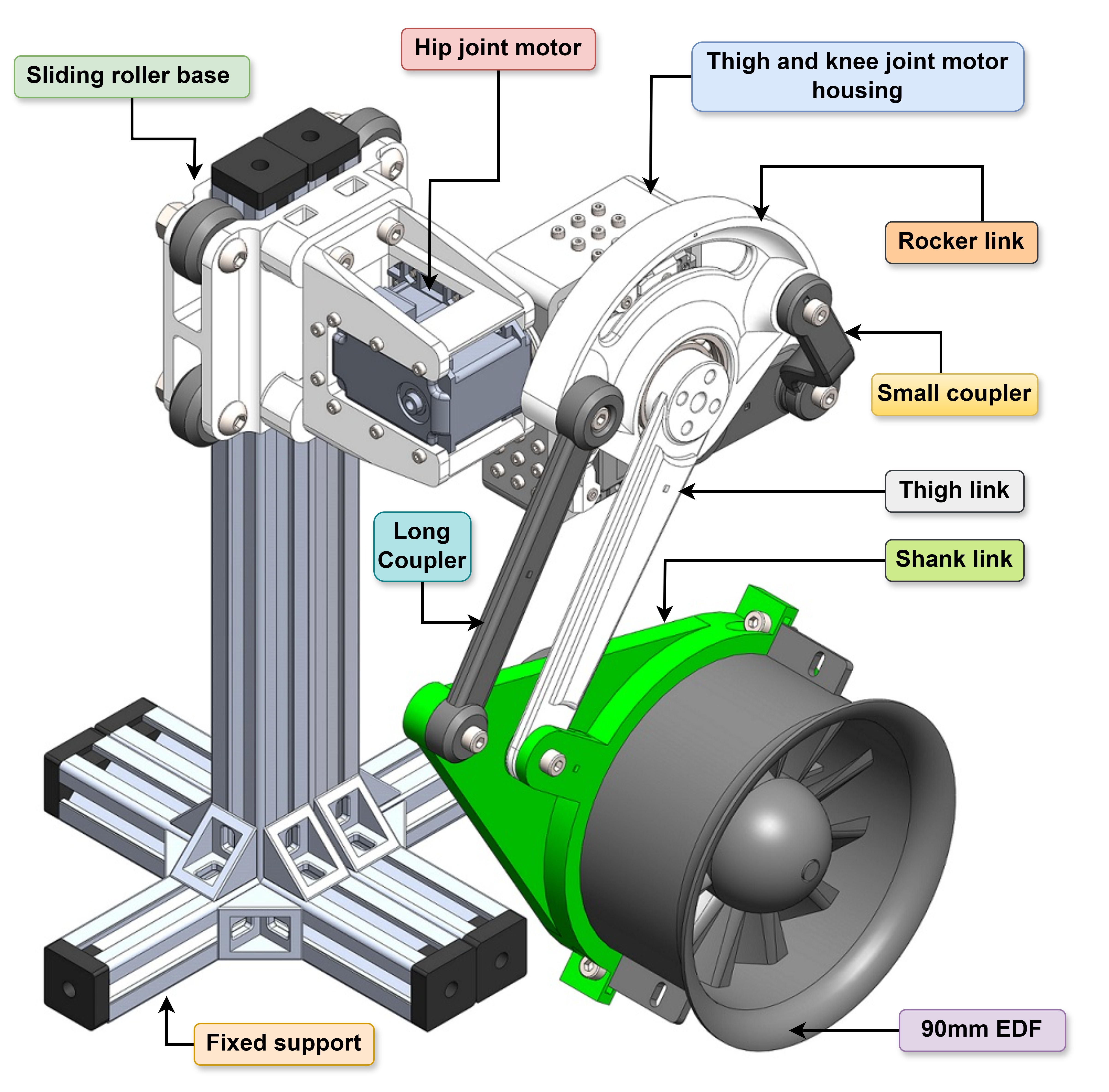}
    \caption{Test rig for the analysis of a single locomotor module}
    \label{rig}
\end{figure}

As the proposed system is larger and more complex, it comprises modules such as locomotors, landing gears, a chassis, and the main central rotor unit. All these subsystems play a role in the proper functioning of the overall system. However, among all the modules, the locomotor module is vital and responsible for the primary actuation of the system for both ground and aerial locomotion. Therefore, in order to understand the locomotor system thoroughly, this work explores the different aspects of this subsystem.\\ 
Fig. \ref{rig} shows a test rig made for the analysis of the locomotor module. Since all locomotion, whether on the ground or in the air, is provided by the locomotor unit. There are the following purposes, which it will solve:
\begin{enumerate}
    \item To understand the full dynamics of individual locomotion.
    \item Gait planning for the quadruped mode implemented in a single leg.
    \item Reconfiguration from the ground mode to aerial mode in a controlled manner.
    \item To test the aerial mode using one arm of the robot.
    \item Based on the insights obtained from the rig, the final version of the locomotor would be attached to the full robot, shown in Fig. \ref{robot_description}.
\end{enumerate}
In the rig, a sliding-roller base component is used in combination with aluminum extrusion profiles. This component actually represents the floating base of the original system. So when the foot is in contact with the ground, then the actuation in the knee and thigh joints will move this sliding element up and down, which will ensure the load-bearing capacity of a single locomotor, as shown in Fig. \ref{rig}. Moreover, the present work is limited to the kinematic aspect only; therefore, the sliding base is locked to a fixed position. However, in the second phase of the overall proposed work, this will be in motion, once the full dynamics of the locomotor is developed.
\subsubsection{Kinematic modeling of the test rig}
\begin{figure}
    \centering
   \begin{minipage}{0.45\textwidth}
    \centering
    \includegraphics[width=\linewidth]{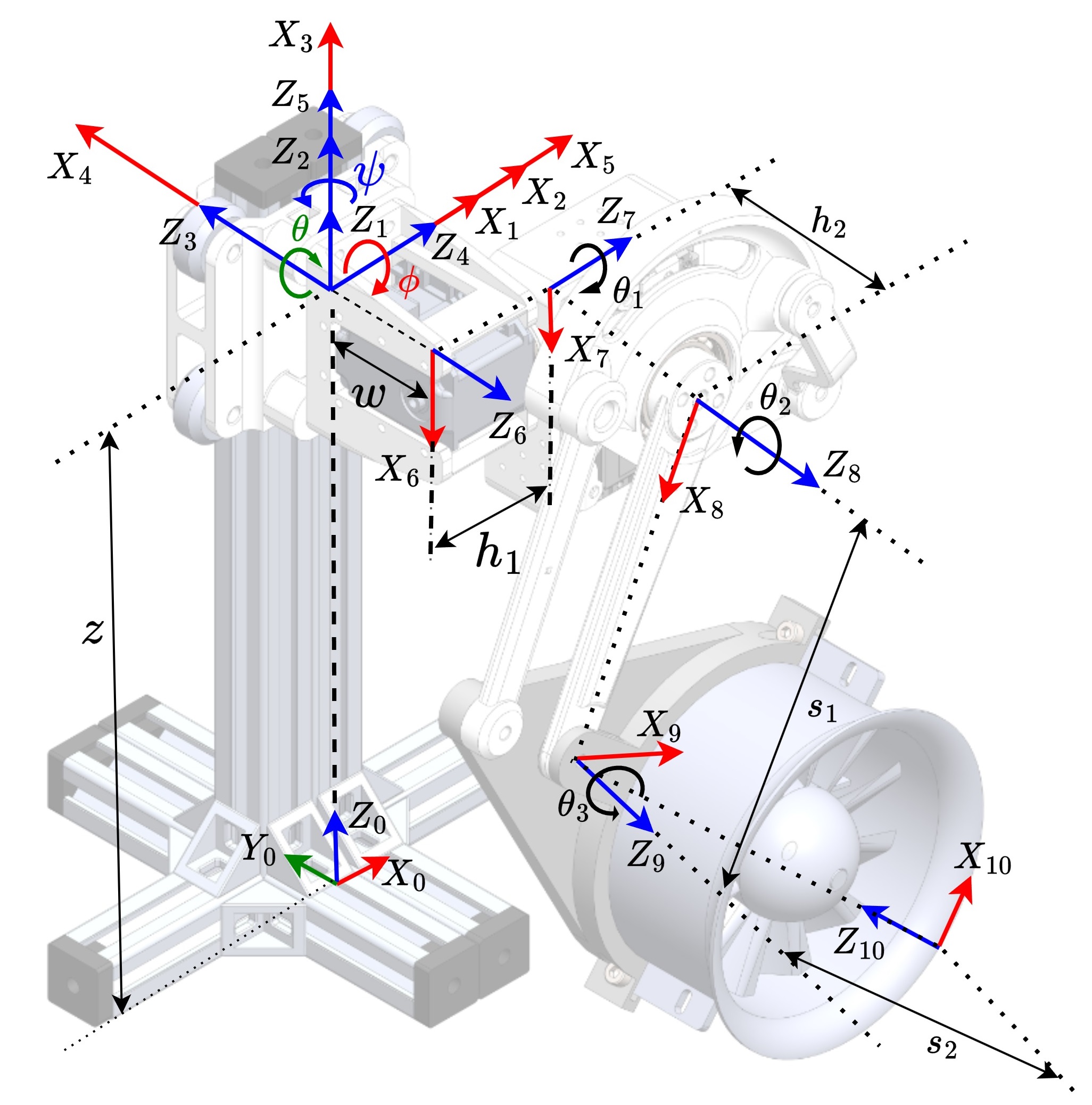}\\
    (a) Parent chain
\end{minipage}
    \begin{minipage}{0.45\textwidth}
        \centering
        \includegraphics[width=\linewidth]{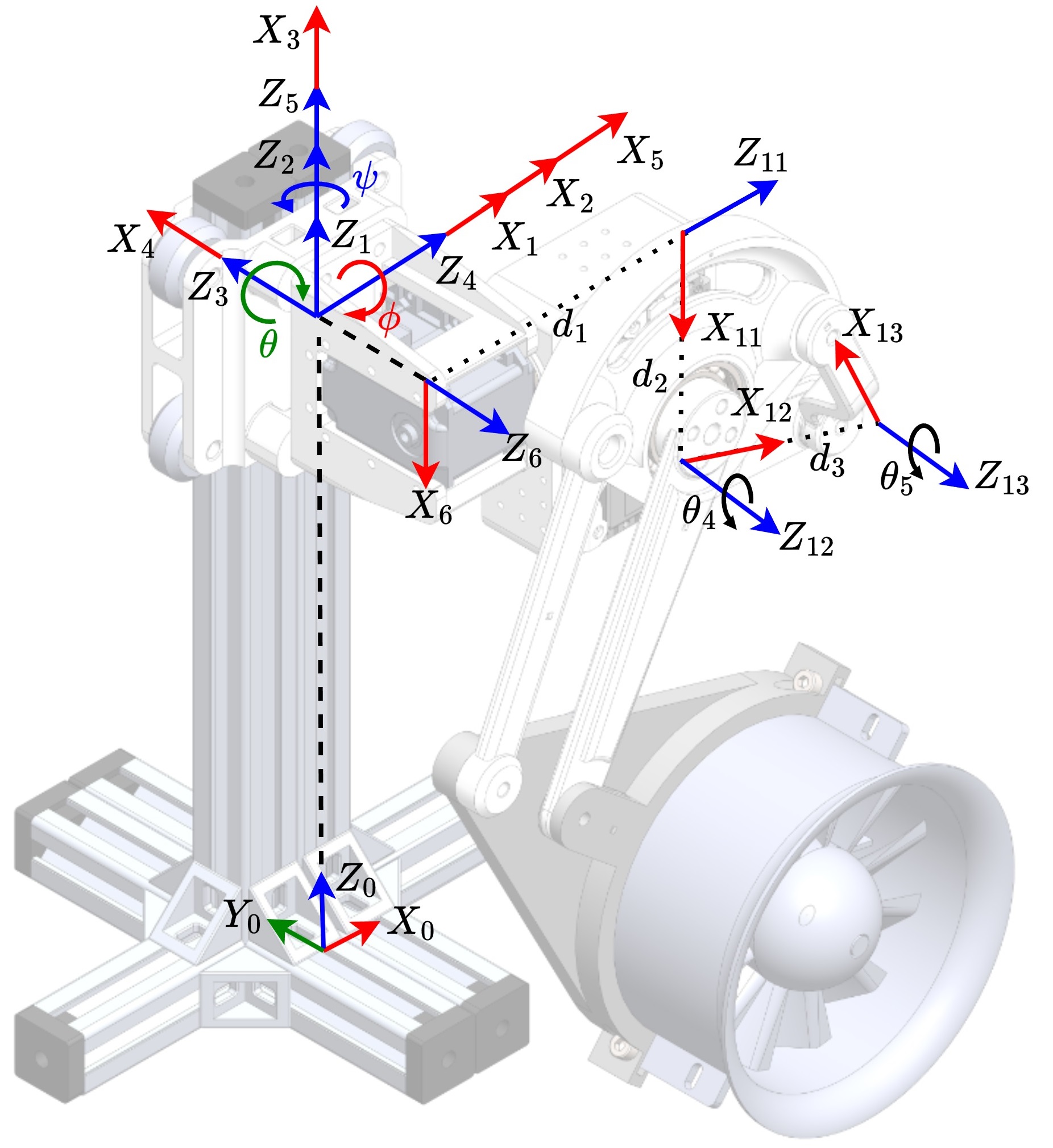}\\
    (b) Small-coupler chain
    \end{minipage}
    \vspace{0.2cm}
    \begin{minipage}{0.45\textwidth}
        \centering
        \includegraphics[width=\linewidth]{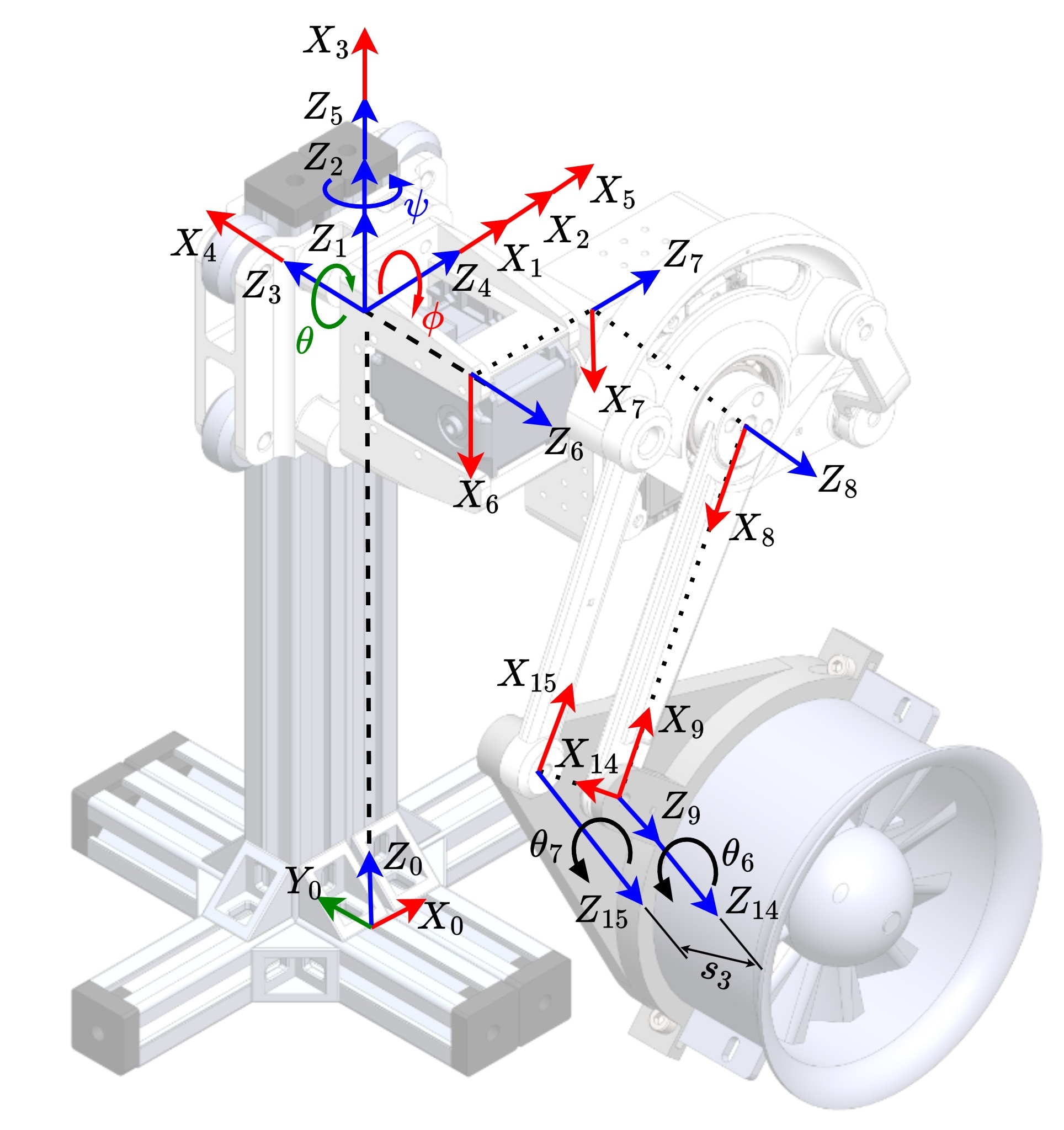}\\
    (c) Long-coupler chain
    \end{minipage}\hfill
    \begin{minipage}{0.45\textwidth}
        \centering
        \includegraphics[width=\linewidth]{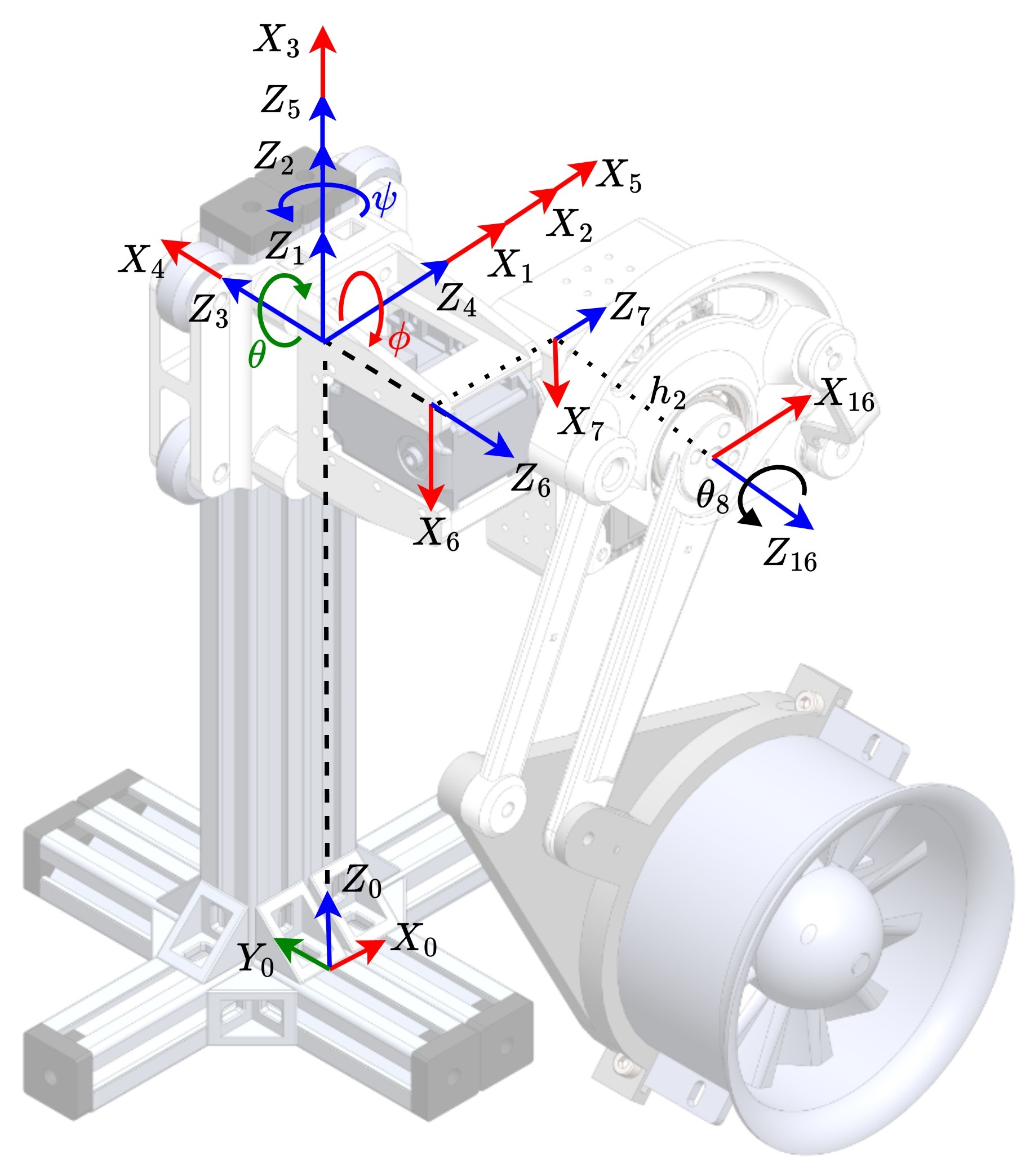}\\
    (d) Rocker chain
    \end{minipage}
    \caption{Kinematic diagrams of different kinematic chains of the test rig.}
    \label{parent_child}
\end{figure}

\begin{figure}
    \centering
    \includegraphics[width=0.6\textwidth]{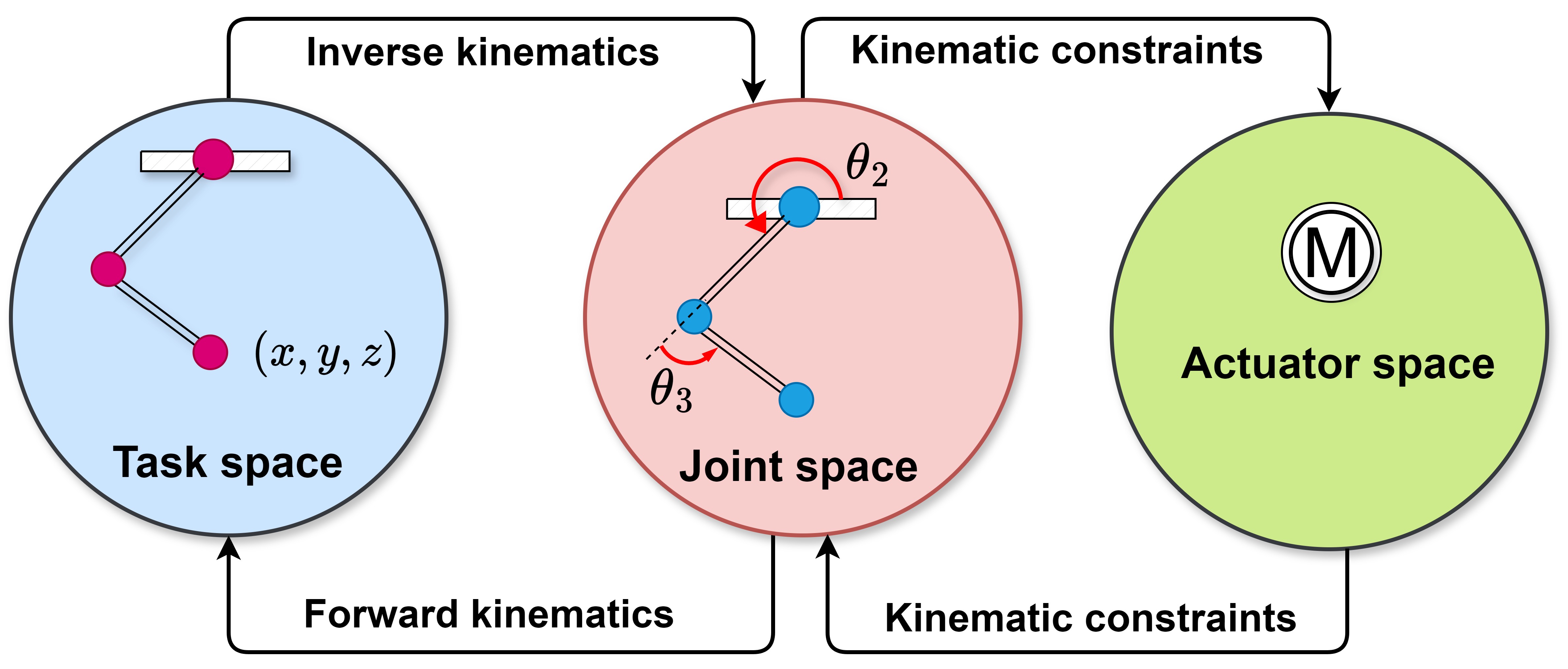}
    \caption{Diagrammatic representation of mapping between different spaces of the rig}
    \label{vector_spac_rig}
\end{figure}

For the analysis of a single locomotor in terms of the motion of the individual link as well as the force experienced at each link and the joints, its kinematic model is developed with the help of a non-standard form of the DH parameter tables. Furthermore, the locomotor is composed of two closed-loop kinematic chains. Since DH parameters are only applicable for open-loop kinematic chains, but due to the presence of closed loops in the kinematic chains, it is not possible to use the same in the original configuration. Therefore, the main configuration of the rig is broken down into the four different open-loop kinematic chains, as shown in Fig. \ref{parent_child}, and their corresponding DH parameter tables are shown in Table \ref{table_3} through Table \ref{table_6}.
The two closed-loop chains in the rig are due to the location of the knee actuator near the floating base. Since the knee joint is primarily away from the thigh and hip joints, which, during abduction/adduction and thigh flexion/extension, create extra load on the hip and thigh joint motors. Secondly, the proximity of the knee actuator to the floating base ensures a higher mass concentration near the robot's body frame, which consequently supports more stable locomotion in both quadruped and airborne modes.
\begin{table*}[t]
\centering

\begin{minipage}[t]{0.48\textwidth}
\centering
\caption{DH parameters: Parent chain.}
\label{table_3}
\renewcommand{\arraystretch}{1.2}
\begin{tabular}{ccccc}
\hline
Frame $i$ & $\alpha_{i-1}$ & $a_{i-1}$ & $\theta_i$ & $d_i$ \\
\hline
1  & 0 & 0 & 0 & $z$ \\
2  & 0 & 0 & $\psi$ & 0 \\
3  & $3\pi/2$ & 0 & $3\pi/2+\theta$ & 0 \\
4  & $3\pi/2$ & 0 & $3\pi/2+\phi$ & 0 \\
5  & $3\pi/2$ & 0 & $3\pi/2$ & 0 \\
6  & $\pi/2$ & 0 & $3\pi/2$ & $w$ \\
7  & $3\pi/2$ & 0 & $\theta_1$ & $h_1$ \\
8  & $\pi/2$ & 0 & $\theta_2$ & $h_2$ \\
9  & 0 & $s_1$ & $\pi/2+\theta_3$ & 0 \\
10 & $3\pi/2$ & 0 & 0 & $-s_2$ \\
\hline
\end{tabular}
\end{minipage}
\hfill
\begin{minipage}[t]{0.48\textwidth}
\centering
\caption{DH parameters: Small coupler.}
\label{table_4}
\renewcommand{\arraystretch}{1.2}
\begin{tabular}{ccccc}
\hline
Frame $i$ & $\alpha_{i-1}$ & $a_{i-1}$ & $\theta_i$ & $d_i$ \\
\hline
1  & 0 & 0 & 0 & $z$ \\
2  & 0 & 0 & $\psi$ & 0 \\
3  & $3\pi/2$ & 0 & $3\pi/2+\theta$ & 0 \\
4  & $3\pi/2$ & 0 & $3\pi/2+\phi$ & 0 \\
5  & $3\pi/2$ & 0 & $3\pi/2$ & 0 \\
6  & $\pi/2$ & 0 & $3\pi/2$ & $w$ \\
11 & $3\pi/2$ & 0 & $\theta_1$ & $d_1$ \\
12 & $\pi/2$ & $d_2$ & $\pi/2+\theta_4$ & 0 \\
13 & 0 & $d_3$ & $\theta_5$ & 0 \\
\hline
\end{tabular}
\end{minipage}

\vspace{0.7cm}

\begin{minipage}[t]{0.48\textwidth}
\centering
\caption{DH parameters: Long coupler.}
\label{table_5}
\renewcommand{\arraystretch}{1.2}
\begin{tabular}{ccccc}
\hline
Frame $i$ & $\alpha_{i-1}$ & $a_{i-1}$ & $\theta_i$ & $d_i$ \\
\hline
1  & 0 & 0 & 0 & $z$ \\
2  & 0 & 0 & $\psi$ & 0 \\
3  & $3\pi/2$ & 0 & $3\pi/2+\theta$ & 0 \\
4  & $3\pi/2$ & 0 & $3\pi/2+\phi$ & 0 \\
5  & $3\pi/2$ & 0 & $3\pi/2$ & 0 \\
6  & $\pi/2$ & 0 & $3\pi/2$ & $w$ \\
7  & $3\pi/2$ & 0 & $\theta_1$ & $h_1$ \\
8  & $\pi/2$ & 0 & $\theta_2$ & $h_2$ \\
9  & 0 & $s_1$ & $\pi/2+\theta_3$ & 0 \\
14 & 0 & 0 & $\pi/2+\theta_6$ & 0 \\
15 & 0 & $s_3$ & $3\pi/2+\theta_7$ & 0 \\
\hline
\end{tabular}
\end{minipage}
\hfill
\begin{minipage}[t]{0.48\textwidth}
\centering
\caption{DH parameters: Rocker.}
\label{table_6}
\renewcommand{\arraystretch}{1.2}
\begin{tabular}{ccccc}
\hline
Frame $i$ & $\alpha_{i-1}$ & $a_{i-1}$ & $\theta_i$ & $d_i$ \\
\hline
1  & 0 & 0 & 0 & $z$ \\
2  & 0 & 0 & $\psi$ & 0 \\
3  & $3\pi/2$ & 0 & $3\pi/2+\theta$ & 0 \\
4  & $3\pi/2$ & 0 & $3\pi/2+\phi$ & 0 \\
5  & $3\pi/2$ & 0 & $3\pi/2$ & 0 \\
6  & $\pi/2$ & 0 & $3\pi/2$ & $w$ \\
7  & $3\pi/2$ & 0 & $\theta_1$ & $h_1$ \\
16 & $\pi/2$ & 0 & $\pi/2+\theta_8$ & $h_2$ \\
\hline
\end{tabular}
\end{minipage}

\end{table*}
The above explanation clearly outlines the rationale behind selecting two closed-loop chains in the leg. Moreover, in future work, for the dynamic analysis of the locomotor, the Newton-Euler recursive algorithm \cite{r25} is selected. However, if the algorithm is applied directly to the main kinematic chain shown in Fig. \ref{parent_child}a, it will calculate the required torque at the knee joint. However, the actuator is not at the knee joint; instead, it is located away from it and transmits the required torque via two closed-loop kinematic chains. Moreover, forces and moments acting on the other links during motion are also required for the design optimization of the different links in the chain. Now, the NE recursive algorithm is applicable only to open-loop kinematic chains natively. Therefore, in order to make it compatible for the dynamic analysis, these two closed loops are broken down into  four kinematic chains, which are illustrated with the help of Fig. \ref{parent_child} and Table \ref{table_3} through Table \ref{table_6}.\\
\begin{figure}
    \centering
    \includegraphics[width=0.55\textwidth]{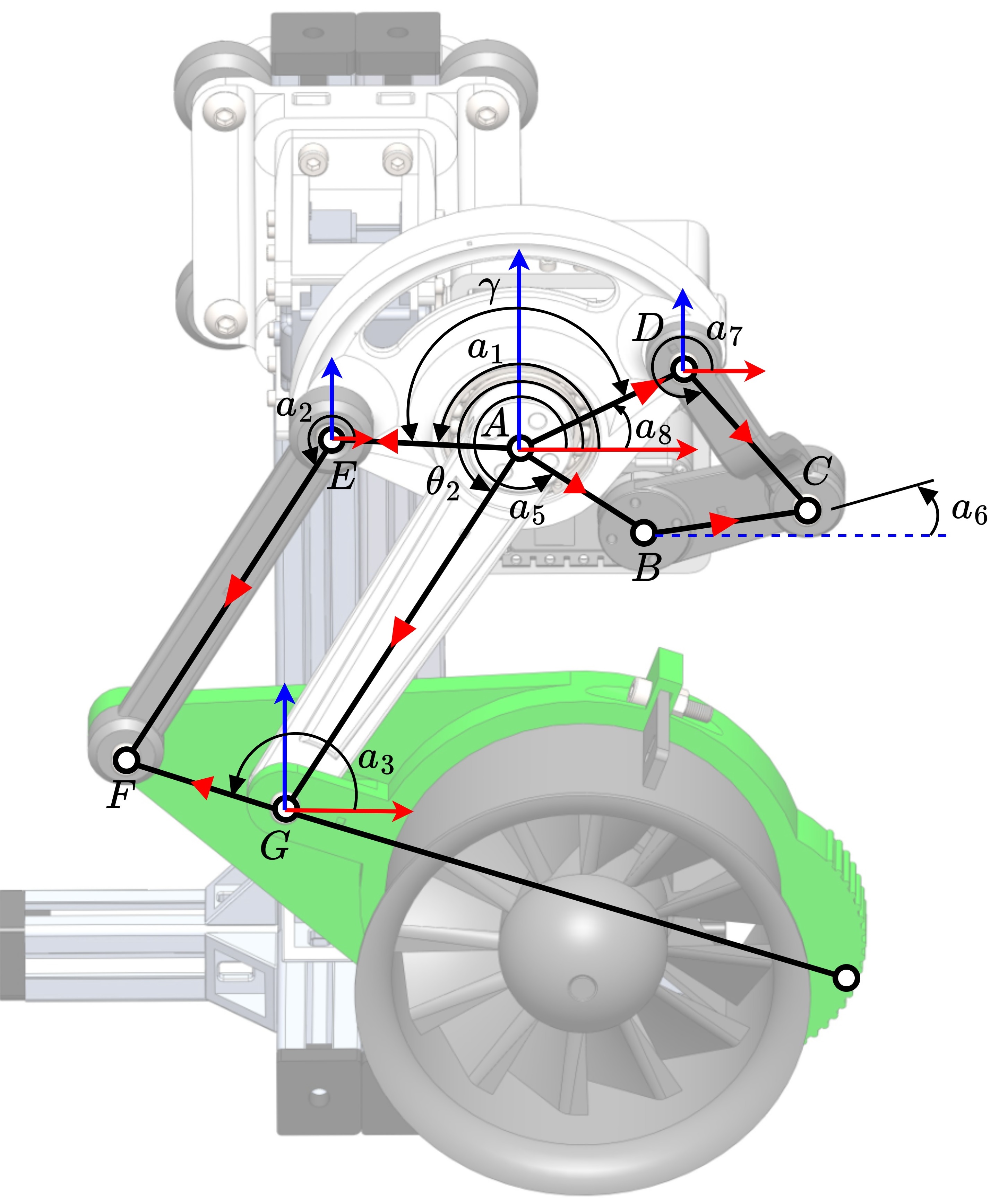}
    \caption{Kinematic diagram of the rig with two closed-loop kinematic chains}
    \label{kin_rig}
\end{figure}
From Fig. \ref{kin_rig}, it can be noted that the knee actuator is located at the point '\textbf{B}', while the actual knee joint is at the point '\textbf{G}'. Therefore, to map the actuator space with the joint space, a mapping function is required; for better understanding, Fig. \ref{vector_spac_rig} can be referred to. Moreover, there are four open-loop kinematic chains for the rig, but on real hardware, these kinematic chains are closed-loop and need to be closed with the appropriate constraints, obtained from the vector loop closure equations \cite{r26} of the closed-loop kinematic chains \textbf{ABCDA} and \textbf{EFGAE}, as represented in equations \eqref{eq_20} and \eqref{eq_21} respectively.
\begin{equation}
    \Vec{AB} + \Vec{BC} - \Vec{CD} - \Vec{DA} = 0
    \label{eq_20}
\end{equation}
\begin{equation}
    \Vec{AE} + \Vec{EF} - \Vec{FG} - \Vec{GH} = 0
    \label{eq_21}
\end{equation}
These two equations are useful for constraining the four open-loop chains and for mapping the actuator space to the joint space. In order to map the actuator space with the joint space equations, \eqref{eq_20} and \eqref{eq_21} are solved simultaneously.
\begin{equation}
    AEe^{ia_{1}}+EFe^{ia_{2}}-GFe^{ia_{3}}-AFe^{ia_{4}}=0
    \label{eq_22}
\end{equation}
\begin{equation}
    ABe^{ia_{5}}+BCe^{ia_{6}}-DCe^{ia_{7}}-ADe^{ia_{8}}=0
    \label{eq_23}
\end{equation}
Equations \eqref{eq_22} and \eqref{eq_23} are the polar forms of the equations \eqref{eq_20} and \eqref{eq_21}, respectively. These two equations are the four equations in Cartesian coordinates with four unknowns. Since, corresponding to the two closed-loop kinematic chains, angles $a_{2}$, $a_{3}$, $a_{7}$, and $a_{8}$ are the unknowns and are dependent on the two actuator angles, $\theta_{2}$ and $a_{6}$. Moreover, out of the remaining two other angles, $a_{1}$ is the function of angle $a_{8}$, and is provided in the appendix \ref{A2}, the second angle is $a_{5}$, which is fixed by the geometry and is also given in the appendix \ref{A2}. It is also noted that the angle $a_{4}$ in the actuator space used in the constraint equation \eqref{eq_22} is equivalent to the angle $\theta_{2}$. Now, there are four equations and four unknowns; therefore, all the unknown angles of the actuator space can be calculated by solving equations \eqref{eq_22} and \eqref{eq_23}, which are provided below:
\begin{equation}
\left.
\begin{aligned}
a_{2} &= 2\arctan2\!\left(-B_{2}+\sqrt{D_{2}},\,2A_{2}\right)\\
a_{3} &= 2\arctan2\!\left(-B_{1}+\sqrt{D_{1}},\,2A_{1}\right)\\
a_{7} &= 2\arctan2\!\left(-B_{3}+\sqrt{D_{3}},\,2A_{3}\right)\\
a_{8} &= 2\arctan2\!\left(-B_{4}+\sqrt{D_{4}},\,2A_{4}\right)\\
\end{aligned}
\right\}
\label{eq_24}
\end{equation}
Where the expression for the terms like, $A_{1}...A_{4}$, $B_{1}...B_{4}$ and $D_{1}...D_{4}$ used in  \eqref{eq_24}, are provided in appendix \ref{A11} and appendix \ref{A12} for brevity.
\section{Jacobian of the leg}
In the previous section, an attempt was made to establish the relationship between the joint space and the actuator space vector using loop closure equations. This section explores the mapping between the task and joint spaces. For that, it is required to have the Jacobian of the entire leg, which accounts for all three spaces: the task space, joint space, and actuator space. The Jacobian of the whole leg can be written as:
\begin{equation}
    \mathbf{V}_{\textbf{f}}= \begin{bmatrix}
        \mathbf{v}_{\textbf{foot}}\\ \bm{\omega}_{\textbf{foot}}
    \end{bmatrix} = \textbf{J}_{\text{leg}}\dot{\textbf{q}}_{a}
    \label{eq_25}
\end{equation}
Where, $\mathbf{V}_{\textbf{f}} \in \mathbb{R}^{6 \times 3}$ is the vector of linear and angular velocity of the foot, $\Dot{\textbf{q}_{a}} \in \mathbb{R}^{3 \times 1}$ is the vector of commanded joint velocities, and, $\mathbf{J}_{\text{leg}} \in \mathbb{R}^{6 \times 3}$ is the Jacobian of the leg, which maps the linear and angular velocity vector of the leg to the commanded joint velocity vector. Now, since the proposed leg consists of two closed-loop kinematic chains for controlling the knee joint. Additionally, these two closed-loop kinematic chains are responsible for enforcing kinematic constraints on all four open-loop kinematic chains, as shown in Fig. \ref{parent_child}. Therefore, in the Jacobian of the leg ($\mathbf{J}_{\text{leg}}$), there must be the contribution of all four open-loop kinematic chains. Now, in order to develop the full Jacobian for the leg, recall the constraint equation in the component form as shown below:
\begin{equation}
    \phi(q) = \begin{bmatrix}
        AE\cos{a_{1}}+EF\cos{a_{2}}-GF\cos{a_{3}}-AF\cos{a_{4}}\\
        AE\sin{a_{1}}+EF\sin{a_{2}}-GF\sin{a_{3}}-AF\sin{a_{4}}\\
        AB\cos{a_{5}}+BC\cos{a_{6}}-DC\cos{a_{7}}-AD\cos{a_{8}}\\
        AB\sin{a_{5}}+BC\sin{a_{6}}-DC\sin{a_{7}}-AD\sin{a_{8}}
    \end{bmatrix}
    \label{eq_26}
\end{equation}
On differentiating \eqref{eq_26} with respect to time:
\begin{equation}
    \Dot{\phi(q)}=\mathbf{J}_{\text{c}}(a)\Dot{\textbf{a}}=0
    \label{eq_27}
\end{equation}
Where, $\mathbf{J}_{\text{c}} \in \mathbb{R}^{4 \times 8}$ is the constraint Jacobian at the velocity level in the actuator space, and $\textbf{a} \in \mathbb{R}^{8}$ is the vector of joint angles ($a=[a_{1}, a_{2}, a_{3}, a_{4}, a_{5}, a_{6}, a_{7}, a_{8}]^{\text{T}}$) in actuator space included in the double closed loop four-bar chains. Out of the eight angles of the actuator space, some are the input angles, and others are enslaved by the geometry. Therefore, the actuator space vector can be rewritten as
\begin{equation}
    \textbf{a} = \begin{bmatrix}
        \textbf{a}_{\text{i}}\\ \textbf{a}_{\text{d}}
    \end{bmatrix}
    \label{eq_28}
\end{equation}
where $\textbf{a}_{i}=\{a_{4}, a_{6}\}$, are the independent angles, and $\textbf{a}_{d}=\{a_{2}, a_{3}, a_{7}, a_{8}\}$, are the dependent angles in the actuator space. Consequently, columns of the $\mathbf{J}_{\text{c}}$ can be reordered as
\begin{equation}
    \mathbf{J}_{\text{c}}=[\mathbf{J}_{\text{ci}}, \mathbf{J}_{\text{cd}}]
    \label{eq_29}
\end{equation}
where $\mathbf{J}_{\text{ci}} \in \mathbb{R}^{4 \times 2}$ and $\mathbf{J}_{\text{cd}} \in \mathbb{R}^{4 \times 4}$ are the Jacobians corresponding to independent joints and the dependent joints of the closed-loop chain, respectively. Therefore, using these two Jacobians, the constraint equation \eqref{eq_27} can be written as:
\begin{equation}
    \mathbf{J}_{\text{ci}}\Dot{\textbf{a}_{\text{i}}} + \mathbf{J}_{\text{cd}}\Dot{\textbf{a}_{\text{d}}}=0
    \label{eq_30}
\end{equation}
If the closed-loop chain is not at singularity and $\mathbf{J}_{\text{cd}}$ is invertible, then:
\begin{equation}
    \Dot{\textbf{a}_{\text{d}}}=-\mathbf{J}_{\text{cd}}^{-1}\mathbf{J}_{\text{ci}}\Dot{\textbf{a}_{\text{i}}}
    \label{eq_31}
\end{equation}
\begin{equation}
    \Dot{\textbf{a}_{\text{d}}} = \textbf{T}_{\textbf{loop}}\Dot{\textbf{a}_{\text{i}}}
    \label{eq_32}
\end{equation}
Where, $\textbf{T}_{\textbf{loop}} = -J_{\text{cd}}^{-1}J_{\text{ci}}$ and $\textbf{T}_{\textbf{loop}} \in \mathbb{R}^{4 \times 2}$, may be defined as the transmission matrix. It indicates how quickly each passive link in the mechanism rotates relative to the input link. It also describes how the motion of the mechanism's links is amplified or blocked in the toggle positions \cite{r26}. In the present case, the actual knee joint angle $\theta_{3}$ is not actuated directly. It is obtained as a geometric consequence of the two closed-loop four-bar chains. Therefore, mathematically, this angle can be represented in the form of a function as:
\begin{equation}
    \theta_{3}=f(a_{6}, \theta_{2})
    \label{eq_33}
\end{equation}
So the actuated coordinates of the overall leg are:
\begin{equation}
    \textbf{q}_{\text{act}}=[\theta_{1}, \theta_{2}, a_{6}]
    \label{eq_34}
\end{equation}
And the actual joint coordinates used in the parent chain are: 
\begin{equation}
    \mathbf{q}=[\theta_{1}, \theta_{2}, \theta_{3}]
    \label{eq_35}
\end{equation}
Now, $\textbf{a}_{\text{i}}=[\theta_{2}, a_{6}]^{\text{T}}$ and $\textbf{a}_{\text{d}}=[a_{2}, a_{3}, a_{7}, a_{8}]^{\text{T}}$ are the independent and dependent vectors, respectively, in \eqref{eq_32}. Angle $a_{1}$ is the function of $a_{8}$, and $a_{5}$ is the fixed angle in the kinematic chain, both of these angles are provided in appendix (\ref{A2}). Using the geometry described in the Fig. \ref{kin_rig}, $\theta_{3}$ can be obtained as:
\begin{equation}
    \theta_{3}=a_{3}-\theta_{2}+\pi
    \label{eq_36}
\end{equation}
Which, on differentiating with respect to the time, gives:
\begin{equation}
    \Dot{\theta_{3}}=\Dot{a_{3}}-\Dot{\theta_{2}}
    \label{eq_37}
\end{equation}
The expression for $\Dot{a_{3}}$ can be obtained from \eqref{eq_31}, as follows:
\begin{equation}
    \Dot{a_{3}}=t_{21}\Dot{\theta_{2}}+t_{22}\Dot{a_{6}}
    \label{eq_38}
\end{equation}
where $t_{21}$ and $t_{22}$ are the elements of the $\textbf{T}_{\textbf{loop}}$ matrix, corresponding to the $\Dot{a_{3}}$.
Substituting \eqref{eq_38} in \eqref{eq_37}, we get:
\begin{equation}
    \Dot{\theta_{3}}=(t_{21}-1)\Dot{\theta_{2}}+t_{22}\Dot{a_{6}}
    \label{eq_39}
\end{equation}
Now the transformation between the actual joint velocity vector and the actuated coordinates can be defined as:
\begin{equation}
   \Dot{\textbf{q}}=\textbf{T}_{\textbf{leg}}\Dot{\textbf{q}}_{\text{act}}
   \label{eq_40}
\end{equation}
Where, $\textbf{T}_{\textbf{leg}}$ is the mapping function between the actuator space and the joint space, and can be described as:
\begin{equation}
    \textbf{T}_{\textbf{leg}}=\begin{bmatrix}
        1 & 0 & 0\\ 0 & 1 & 0\\ 0 & (t_{21}-1) & t_{22}
    \end{bmatrix}
    \label{eq_41}
\end{equation}
Also, the mapping between the task space velocities and the joint space velocities of the leg can be written as:
\begin{equation}
    \mathbf{V}_{\mathbf{foot}}=\mathbf{J}_{\text{parent}}(\theta_{1},\theta_{2},\theta_{3})\Dot{\textbf{q}}
    \label{eq_42}
\end{equation}
Where, $\mathbf{J}_{\text{parent}}\in\mathbb{R}^{6 \times 3}$ is the Jacobian between the task space velocities and the joint space velocities, corresponding to the Fig. \ref{parent_child}a and the Table \ref{table_3}.
On substituting \eqref{eq_40} into \eqref{eq_42}, we get:
\begin{equation}
\mathbf{V}_{\mathbf{foot}}=\mathbf{J}_{\text{parent}}\mathbf{T}_{\textbf{leg}}\Dot{\textbf{q}}_{\text{act}}
     \label{eq_43}
\end{equation}
Where, $\mathbf{J}_{\text{parent}}\mathbf{T}_{\textbf{leg}}=\mathbf{J}_{\text{leg}}$, is already defined in \eqref{eq_25}. In this section, the full-leg Jacobian is formulated, which is very helpful for mapping joint and task-space velocities. Also, it is helpful to assess the workspace quality to determine the feasible singularity-free region for leg motion, which will be discussed in the next section.
\section{Workspace and singularity analysis}
\begin{figure}
    \centering
    \includegraphics[width=0.75\linewidth]{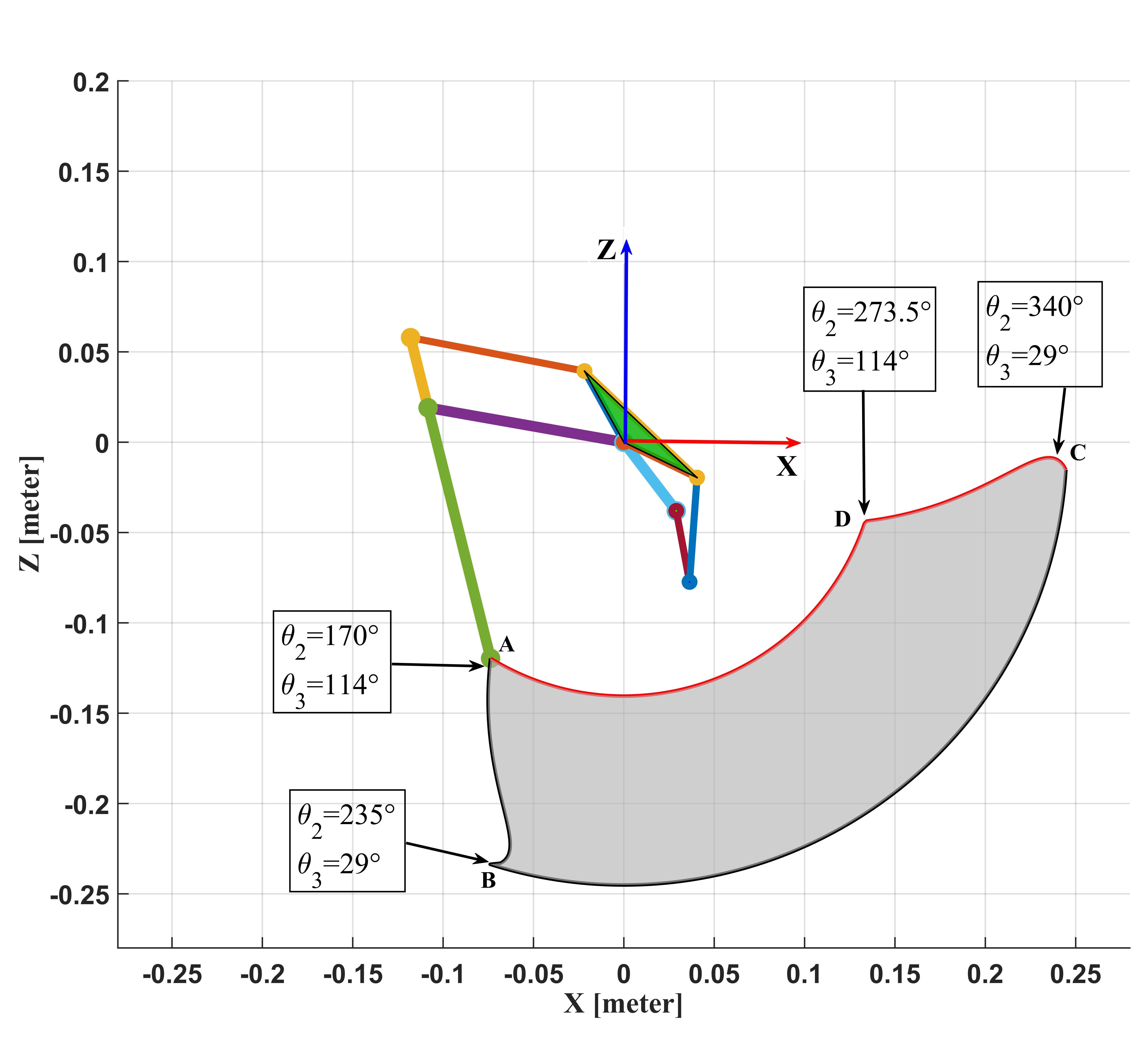}
    \caption{Reachable workspace of the leg in task space}
    \label{task_space}
\end{figure}

The workspace of the leg is a very important aspect of the overall system, since it provides a strong foundation for the leg's path planning under different locomotion requirements. As for the present work, it is planned to generate different motion paths, such as trot gait and crouching gaits, but for that, it is very important to have clear information about the limits of the actuator angles in actuator space or the joint angles in joint space. Therefore, in the present section, a workspace analysis of the rig is carried out to understand the leg's motion in both the Cartesian and joint spaces. Moreover, the workspace is verified using the singularity check throughout the workspace points.
\paragraph{\textbf{Methodology used for the workspace and point cloud generation:}}
To calculate the leg's workspace, the hard limits of all actuators were first examined using SolidWorks 2019. These hard limits are pointed out by the points A, B, C, and D in Fig. \ref{task_space}. Once these joint limits are set, then both the angles $\theta_{2}$ and $\theta_{3}$ are varied with a resolution of $0.5^\circ$, which, however, can be changed to any other value based on the actuator resolution. However, for the current work, the workspace resolution is fixed at $0.5^\circ$. Both joint angles are varied in a particular manner within their hard limits, as discussed in the subsequent section. Once both joints reach all possible limits, the corresponding joint angles and the foot coordinates are saved in the spreadsheet file, which serves as the required point cloud and will be useful for generating gaits. All the gaits generated through this point cloud are discussed in the results and discussion section. In this section, both the joint and task spaces are discussed, along with a singularity check for all points in the workspace.
\subsection{Workspace in Cartesian coordinates}
In order to find out the workspace in task space, only the two angles, $\theta_{2}$ and $\theta_{3}$ are used. With the inclusion of these two angles, the workspace will be a plane only, as shown in Fig. \ref{task_space}. However, if angle $\theta_{1}$ is also included, the workspace will be a volume. But since $\theta_{1}$ is independent of the contribution in the closed loop chains, its variation is not discussed here.
As it can be observed from the Fig. \ref{task_space}, the leg's motion is shown in a 2D plane with only two angles, $\theta_{2}$ and $\theta_{3}$ as the parameters. In the same figure, it is noted that, along the path A-B-C, angle $\theta_{2}$ is varying from $170^\circ$ to $340^\circ$, and angle $\theta_{3}$ is varying from $114^\circ$ to $29^\circ$, which in terms of the legged robot's terminology is the pseudo swing phase. Similarly along the path C-D-A, angle $\theta_{2}$ is varying from $340^\circ$ back to $170^\circ$, and angle $\theta_{3}$, from $29^\circ$ to $114^\circ$, which we can call as the pseudo stance phase. As we can see from the Fig. \ref{task_space} that the swing and stance phase should follow a smooth trajectory, which however, will be planned in the gait generation section later. But here the objective was to generate the reachable workspace and to obtain the corresponding point cloud, which later indeed will be helpful to plan the gaits. Moreover, in the same figure, it is also noted that both the joint angles are connected with each other through the constraint equations \eqref{eq_22} and \eqref{eq_23}. So while the foot is made to move between two points in the workspace, let say between point-A to point-B, both the angles are required to satisfy the closed loop constraints.
\begin{figure}
    \centering

    \begin{minipage}{0.45\textwidth}
        \centering
        \includegraphics[width=\linewidth]{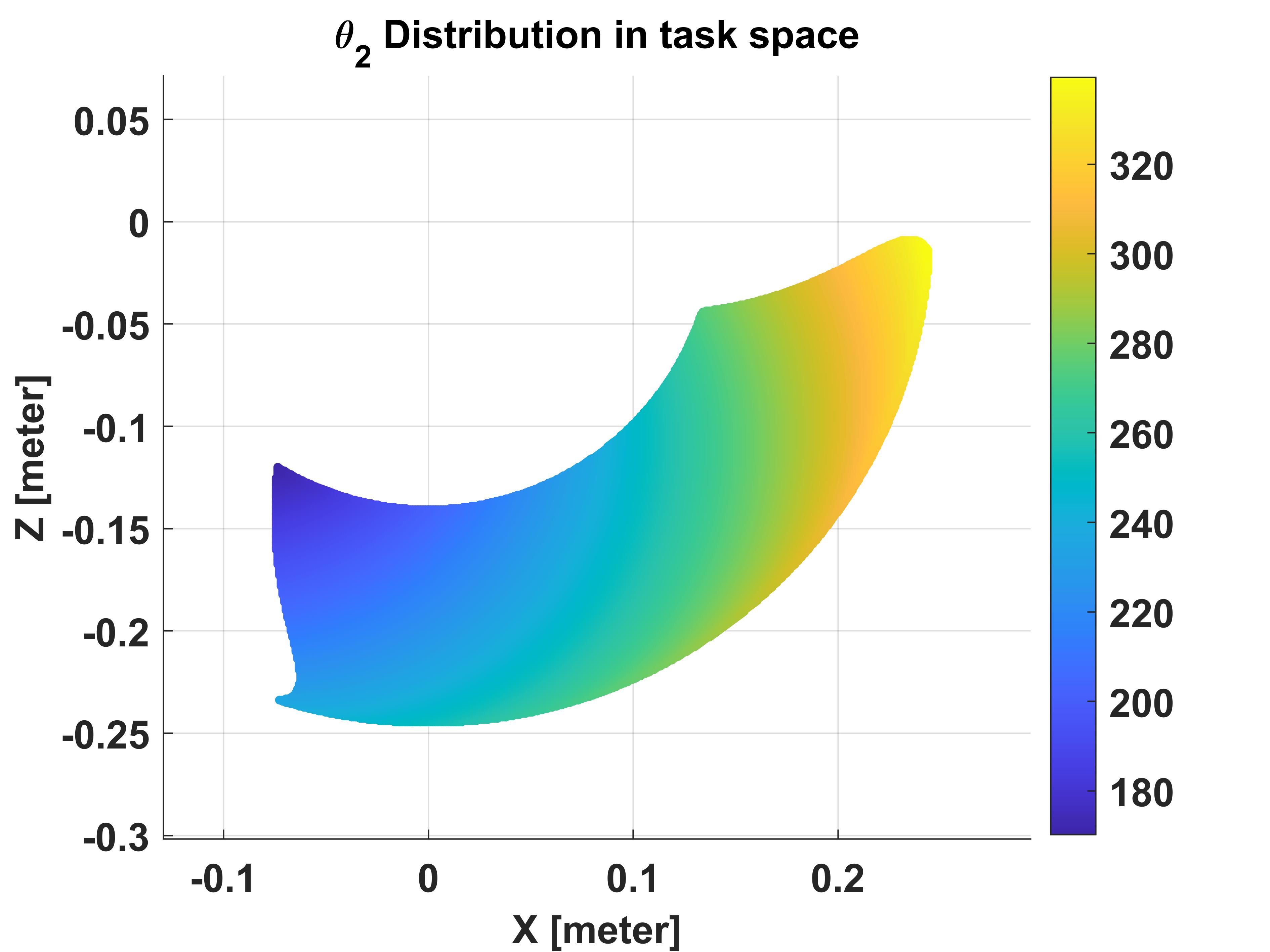}
        \small (a) Task space with variation of angle $\theta_{2}$
    \end{minipage}
    \hfill
    \begin{minipage}{0.45\textwidth}
        \centering
        \includegraphics[width=\linewidth]{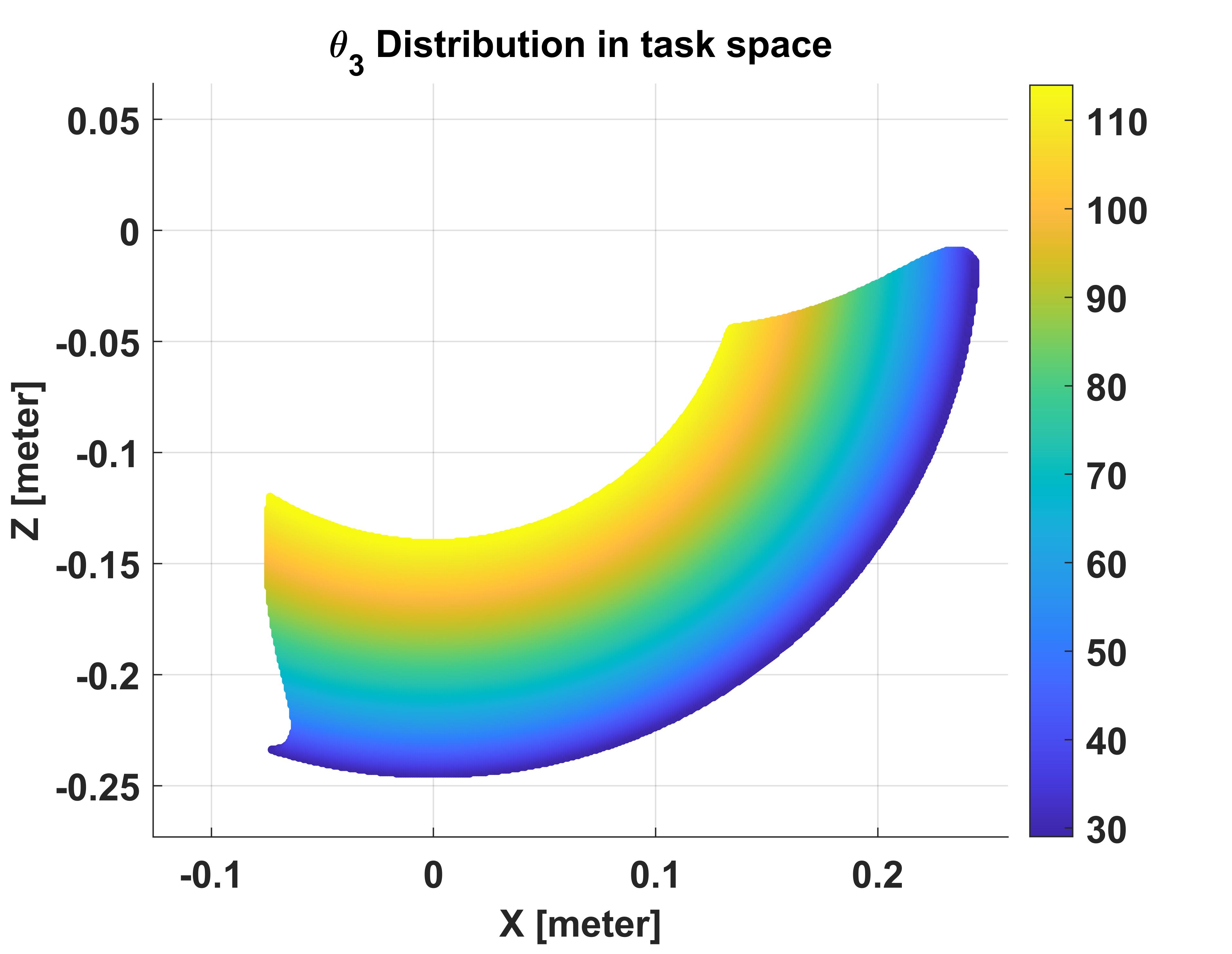}
        \small (b) Task space with variation in angle $\theta_{3}$
    \end{minipage}

    \caption{Distribution of joint angles in the task space.}
    \label{fig_ts}
\end{figure}
Consequently, distribution of the angle $\theta_{2}$ in Fig. \ref{fig_ts}a is different from the distribution of $\theta_{3}$ in Fig. \ref{fig_ts}b.
\subsection{Workspace in joint coordinates}
\begin{figure}
    \centering

    \begin{minipage}{0.45\textwidth}
        \centering
        \includegraphics[width=\linewidth]{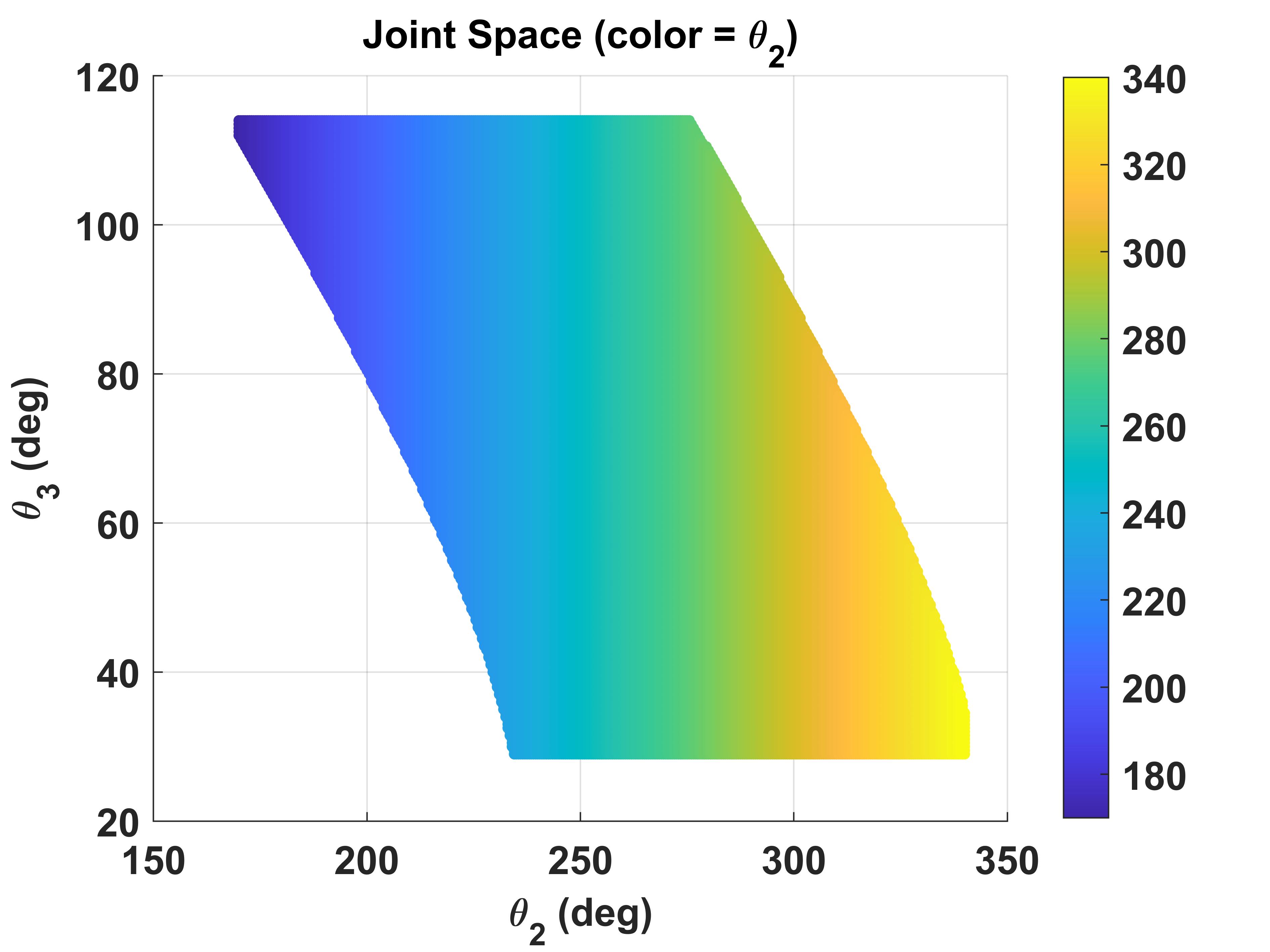}
        \small (a) Joint space with variation of angle $\theta_{2}$
    \end{minipage}
    \hfill
    \begin{minipage}{0.45\textwidth}
        \centering
        \includegraphics[width=\linewidth]{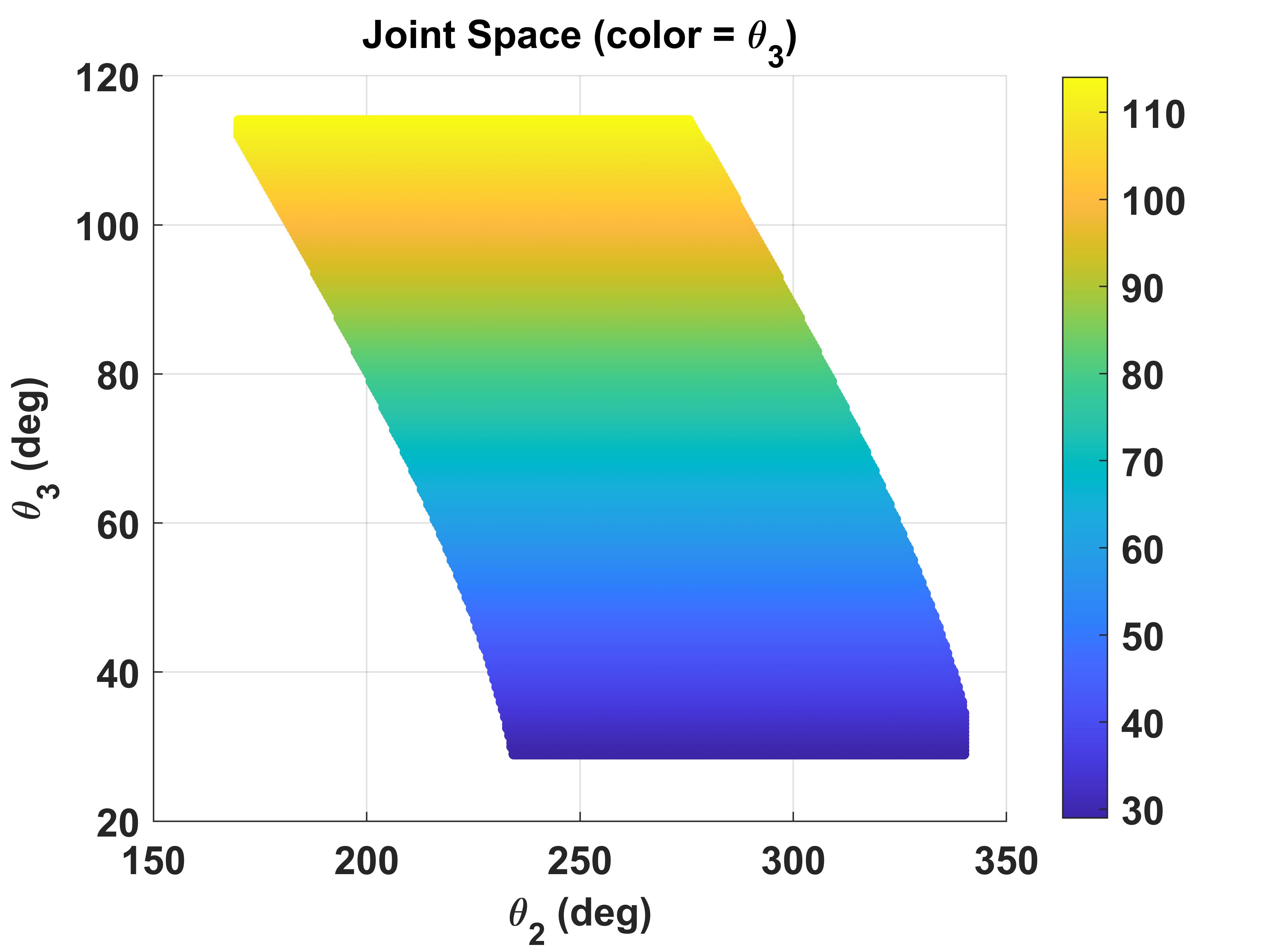}
        \small (b) Joint space with variation in angle $\theta_{3}$
    \end{minipage}

    \caption{Workspace of the leg in joint-space}
    \label{fig_js}
\end{figure}

As it has already been discussed previously, in order to obtain the workspace of the leg, joint angles $\theta_{2}$ and $\theta_{3}$ are varied under the physical joint limits and the closed-loop kinematic constraints. Additionally, in the Jacobian section, the mapping between the task, joint, and actuator spaces was discussed. Therefore, if we observe the Fig. \ref{fig_js}, which is the joint space formed by the angle $\theta_{2}$ and $\theta_{3}$ is the reflection of the foot position of the leg in joint space. And both of these spaces can be obtained from joint-space to task-space using forward kinematics, or from task-space to joint-space using inverse kinematics, as discussed previously.\\
In the proposed work, forward kinematics is used to generate the workspace in joint space. Since the hard limits of the joints were already known from the system's CAD model, the joint angles were mutually related through closed-loop kinematic constraints. 
Additionally it is also worth to note that, corresponding to the joint angles $\theta_{2}$ and $\theta_{3}$, their parent angle in the actuator spaces are the angles $\theta_{2}$ (equivalent to $a_{4})$ and $a_{6}$ respectively, which can be verified from the Fig. \ref{kin_rig}. Now, out of angles $\theta_{2}$ and $a_{6}$, angle $\theta_{2}$ is falling in both joint-space and in the actuator space. However, the joint-space angle $\theta_{3}$ is related to the actuator space angle $a_{6}$ through the closed-loop kinematic constraints. As a result, the leg’s reachable workspace, when represented in actuator space, is not entirely enclosed within that space. Since one actuator is falling into the joint space, and due to which there is partial overlapping between the joint-space and the actuator-space. Consequently, the reachable workspace of the leg in the actuator space is not included in this work. However, in the point cloud, corresponding to every foot coordinate, its corresponding joint-space coordinate and actuator-space coordinates are stored for hardware implementation purposes.
\subsection{Workspace singularity analysis}
\begin{figure}
    \centering
    \includegraphics[width=0.8\linewidth]{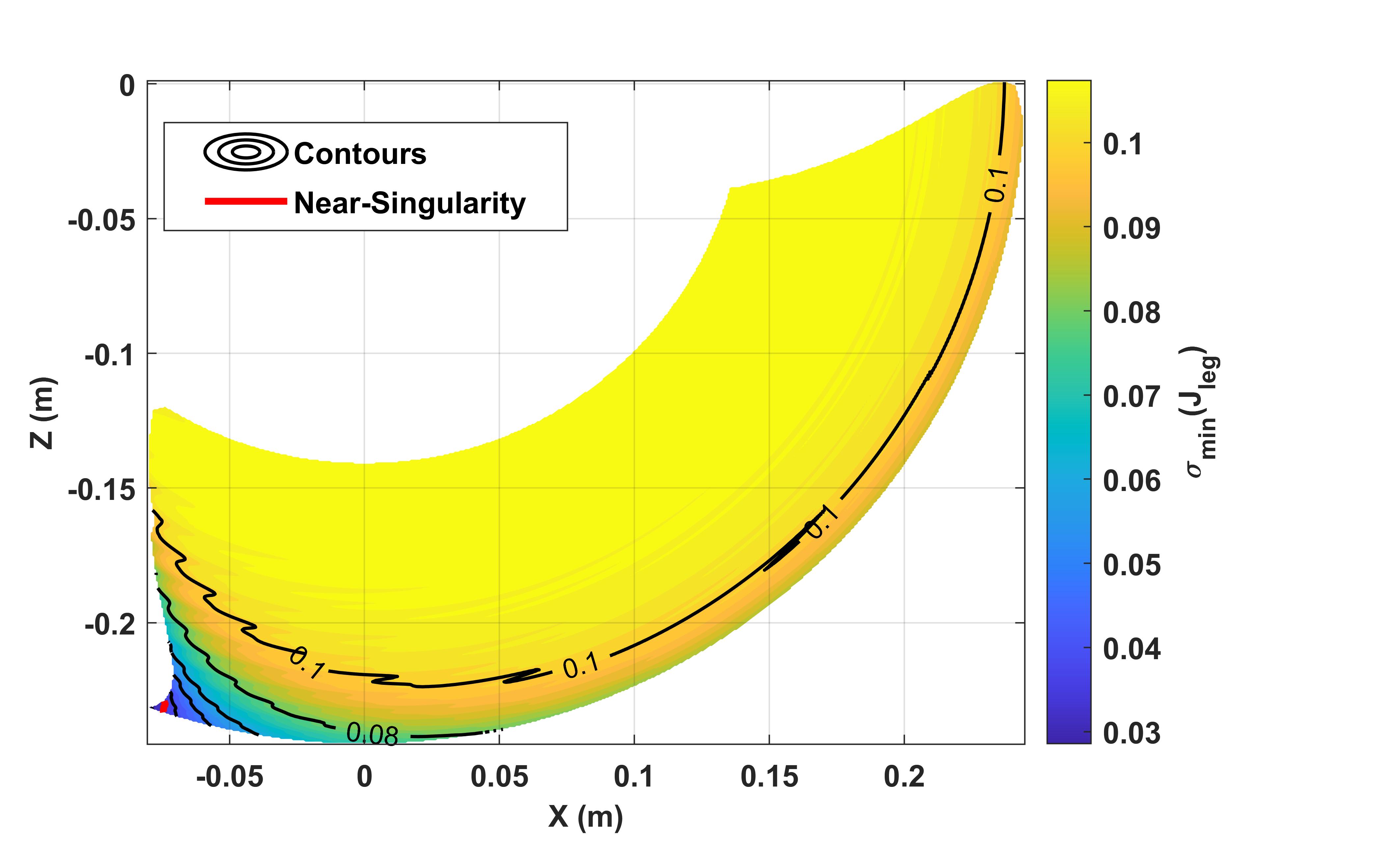}
    \caption{Singularity in the workspace in Cartesian space}
    \label{ts_singularity}
\end{figure}

\begin{figure}
    \centering
    \includegraphics[width=0.7\linewidth]{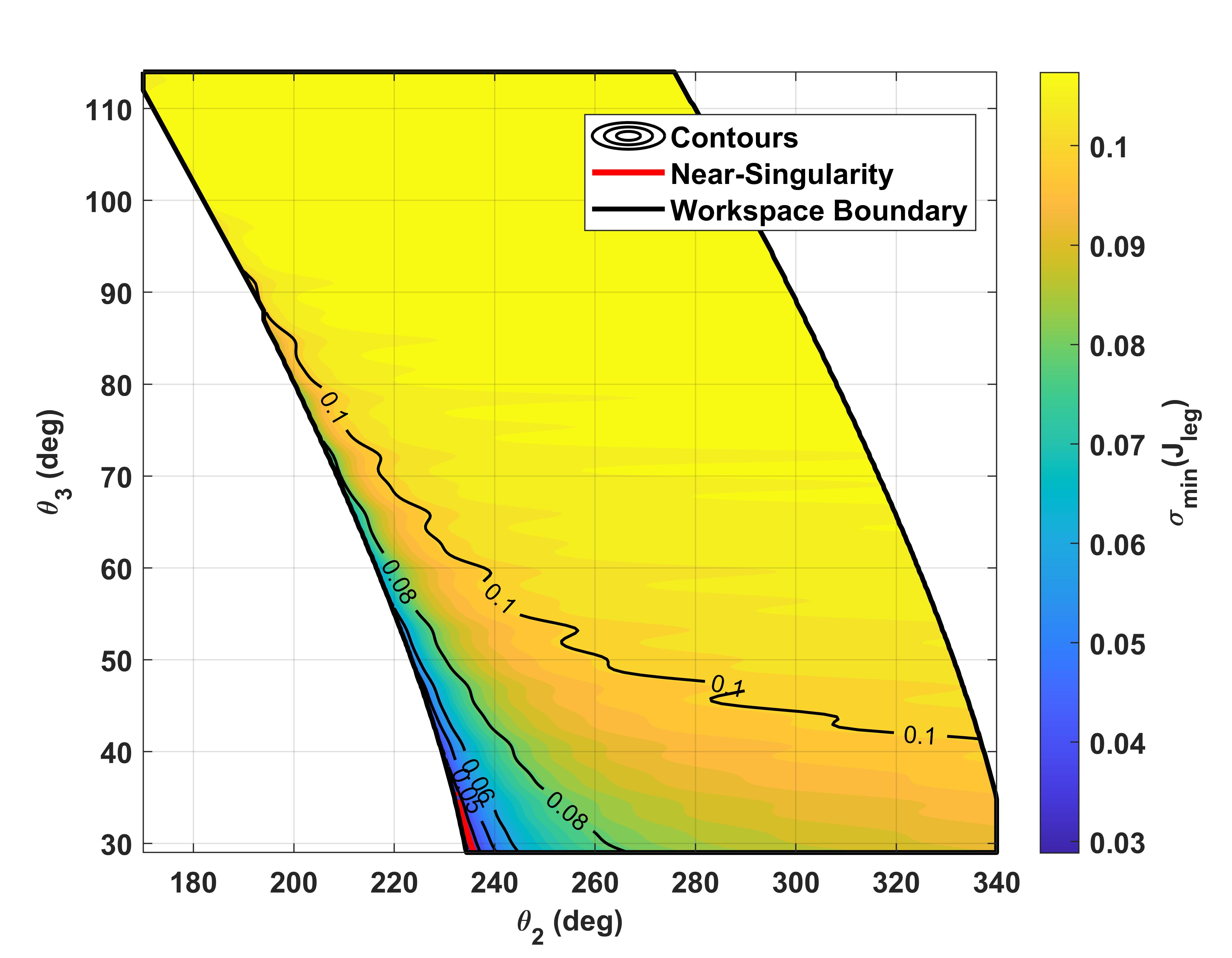}
    \caption{Joint space singularity}
    \label{js_singularity}
\end{figure}

Singularity analysis of the reachable workspace of the proposed locomotor is significant due to the following two reasons:
\begin{itemize}
    \item There are two closed-loop kinematic chains used in the system, in which if the linkages of any of the loops attain the toggle positions, it may lead to an infinite torque required at the actuator, and the corresponding state of the system will be singular.
    \item Secondly, if corresponding to any point in the task space, the Jacobian of the leg ($\mathbf{J}_{leg}$) is ill-conditioned or $det(\mathbf{J}_{leg}) = 0$, then we may not be able to control the foot position and velocity through the actuator.
\end{itemize}
\paragraph{\textbf{Singularity in closed-loop chain:}}The reason for singularity at the kinematic-loop level can be described by the $\mathbf{T_{loop}}$. Since this matrix is the function of the $\mathbf{J_{cd}}$. Therefore, to check the singularity at the loop level, matrix $\mathbf{J_{cd}}$ should not be singular.
\begin{equation}
    \mathbf{J_{cd}} = \begin{bmatrix}
        -EF\sin{a_{2}} & GF\sin{a_{3}} & 0 & 0\\
        EF\cos{a_{2}} & -GF\cos{a_{3}} & 0 & 0\\
        0 & 0 & DC\sin{a_{7}} & AD\sin{a_{8}}\\
        0 & 0 & -DC\cos{a_{7}} & -AD\cos{a_{8}}
    \end{bmatrix}
    \label{eq_44}
\end{equation}
Now on substituting $det(\mathbf{J_{cd}})=0$, we get:
\begin{equation}
\left.
\begin{aligned}
a_{2} - a_{3} &= n\pi, \\
a_{7} - a_{8} &= n\pi,
\end{aligned}
\right\}
\quad \text{where } n = 0, 1, 2, \dots
\label{eq_45}
\end{equation}
From equation \eqref{eq_45}, it can be observed that the matrix $\mathbf{J_{cd}}$ is singular when either or both link pairs, EF-FG and AD-DC, are collinear (Fig. \ref{kin_rig}). However, the link lengths of the leg are chosen to avoid this situation. But there is still some effect on the leg due to the link EF and FG, which can be observed in the Fig. \ref{ts_singularity} and Fig. \ref{js_singularity}. From both of these two figures, it can be seen that there is a small red region corresponding to the joint angle $\theta_{2}=235^\circ$ and $\theta_{3}=29^\circ$. This is because at this joint angle, the links EF and FG are nearly collinear. But due to the hard limits imposed by the link's geometry, this collinearity is avoided safely.\\
\paragraph{\textbf{Singularity in full leg:}}Another reason for the singularity in the leg’s workspace arises from the Jacobian $\mathbf{J_{leg}}$, as illustrated by equation \eqref{eq_25}.
As it can be noted that the $\mathbf{J_{leg}}$ is the function of $\mathbf{J_{parent}}$ (provided in the appendix \ref{A3}) and the $\mathbf{T_{leg}}$ (equation \eqref{eq_41}). So to obtain the singularity-free workspace, $det(\mathbf{J_{leg}})\ neq 0$, throughout the workspace. Therefore, corresponding to every point of the point-cloud obtained earlier and using the singular value decomposition (SVD), $\sigma_{min}$ is calculated and plotted in the task space, Fig. \ref{ts_singularity}, and the joint space, figure(\ref{js_singularity}). From these figures, it can be observed that the $\sigma_{min}$ variation is between 0.03 and 0.1.
In the bottom left part of the Fig. \ref{js_singularity}, it can be observed that the $\sigma_{min}$ contour lines are denser and closer to the workspace boundary. Moreover, in this same region, the value $\sigma_{min}$ is close to zero (0), corresponding to the singular points in the workspace.\\ At this stage, we now have two clear pictures of the cause of singularity in the workspace. One is due to the singularity at the closed-loop kinematic chain level, and the second is due to the Jacobian of the full leg. But if we observe both together, we can see that the singularity at the loop level occurred when any two links of the closed-loop kinematic chain were collinear, or when the configuration reached the toggle position. These toggle positions, particularly in four-bar chains, generally lock the mechanism if driven by the rocker. But in the present case, the driver link is crank, so it can easily come out of the toggle lock. Additionally, neither of these closed-loop four-bar kinematic chains follows the Grashof condition; therefore, this toggle in the present case is avoided by modeling the link in such a way that its geometry will resist the two adjoining links getting collinear. So, this is the very reason for getting singular points at the loop level and from the leg's Jacobian. 
\section{Results and discussion}
\begin{figure}
    \centering
    \includegraphics[width=0.9\linewidth]{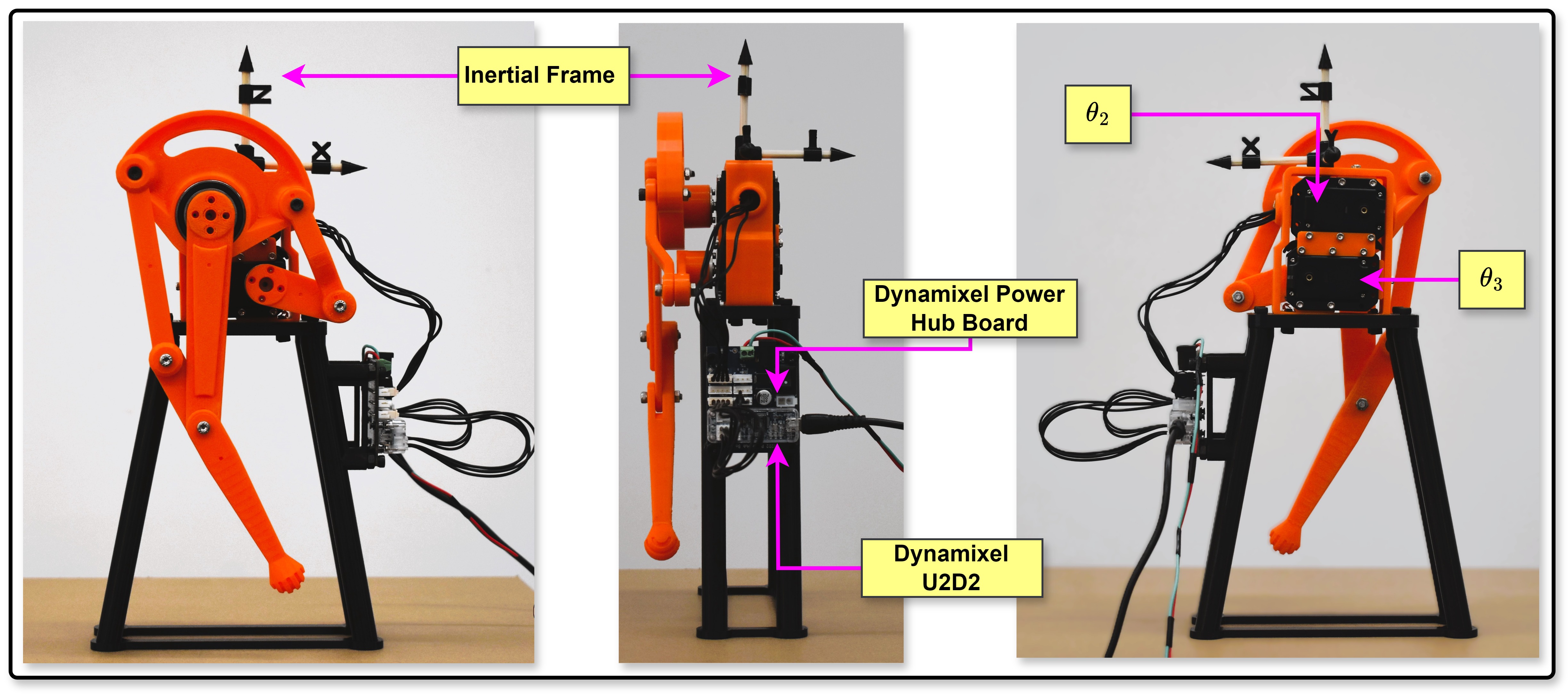}
    \caption{Single-leg test rig for the analysis of workspace}
    \label{small_rig}
\end{figure}

\begin{figure}
    \centering
    \includegraphics[width=0.9\linewidth]{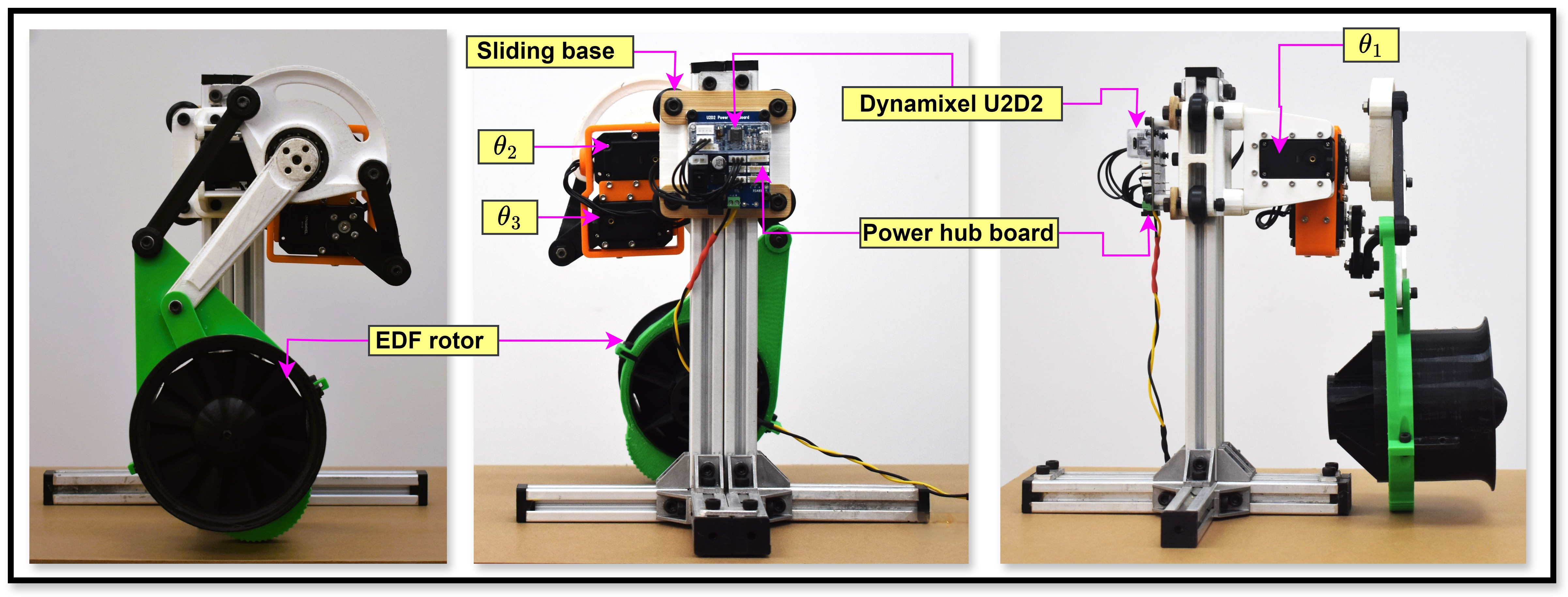}
    \caption{Single-leg test rig with all three joints and EDF}
    \label{big_rig}
\end{figure}

The kinematics of the leg is established with the full workspace and singularity analysis. In this section, the kinematic implementation on real hardware is discussed.
\paragraph{\textbf{Hardware description:}} In order to verify the motion or kinematics of the leg, a test bench is created. This test bench consists of a single leg only. All the links of the legs are selected to satisfy the non-Grashof condition required by the proposed leg and are verified in AutoCAD 2024. Later, based on the verified lengths, all links are modeled in SolidWorks 2019. 
\begin{figure}
    \centering
    \includegraphics[width=0.65\linewidth]{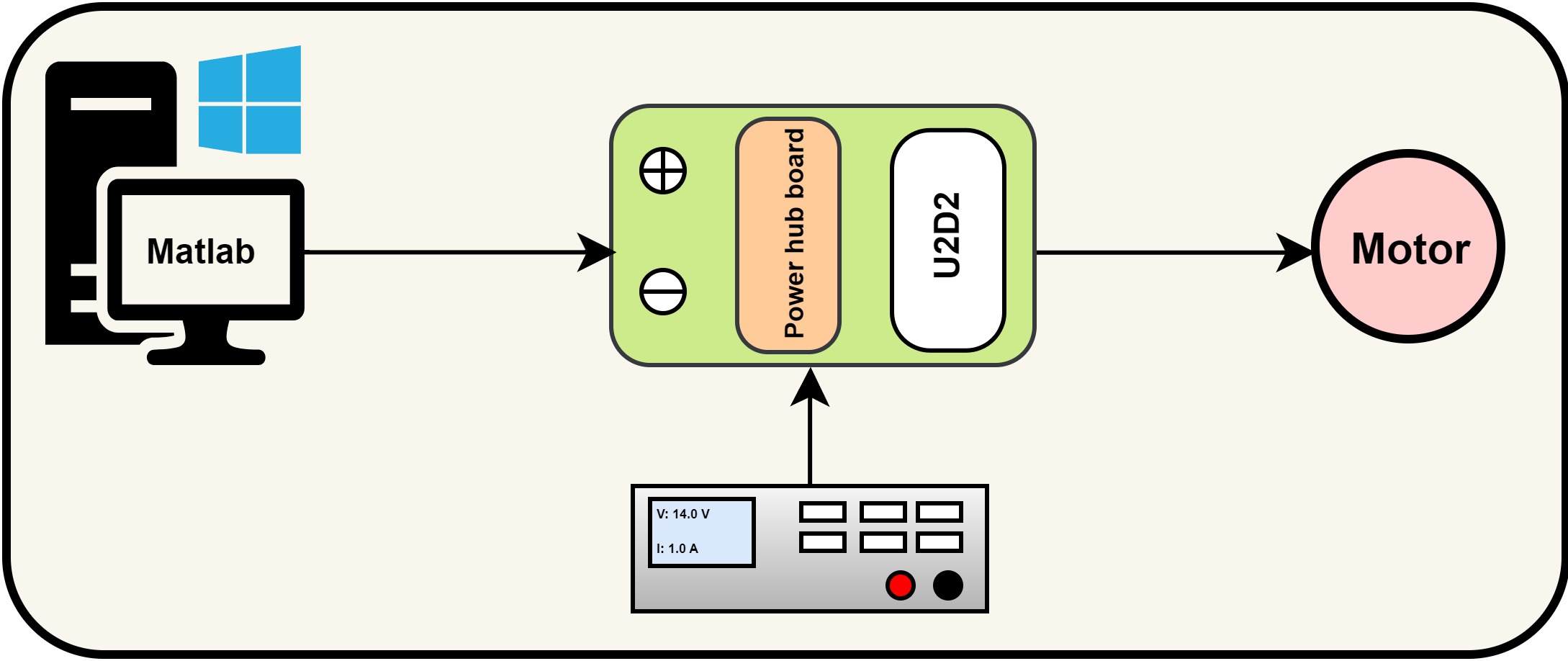}
    \caption{Schematic diagram of the communication between the computer and the experimental setup}
    \label{U2D2}
\end{figure}
All the links are 3D printed using PLA and carbon fiber composite filaments. For joint actuation, Dynamixel MX28T motors are used, along with their own U2D2 for serial communication with the computer and a power hub board for powering the motors. The overview of communication between the computer program and the actuator is easily understood from Fig. \ref{U2D2}. All three actuators use TTL communication and are connected in a daisy chain with unique individual IDs, all at the same baud rate.\\ For conducting the experiment, to validate the workspace and other simple gaits, two test benches are created. The first test bench, as shown in Fig. \ref{small_rig}, consists of only two actuators. Out of these two, the first actuator corresponds to the joint angle $\theta_{2}$ and the second to $\theta_{3}$. This test bench is designed to study workspace analysis and leg gait planning, which only require the joint motors for $\theta_{2}$ and $\theta_{3}$. These two joint motors are annotated by $\theta_{2}$ and $\theta_{3}$ in the picture. But since the overall system is a multimodal type and requires reconfiguration of the locomotor module, as discussed in the introduction, a second test bench is fabricated, as shown in Fig. \ref{big_rig}. This test bench is equipped with a joint motor $\theta_{1}$ along with two other motors. This motor serves two purposes: first, it provides abduction and adduction at the hip joint for the leg mode of the system; second, it helps the leg change orientation, which is required for the aerial mode. In aerial mode, a 90 mm EDF unit is also attached to the end of the leg. However, this EDF is inactive for now, since the scope of the present work is limited to the kinematics only. The remaining hardware and electrical connection scheme of this test bench is the same as the previous one.
\subsection{Validation of the reachable workspace of the leg}
\begin{figure}
    \centering
    \includegraphics[width=0.9\linewidth]{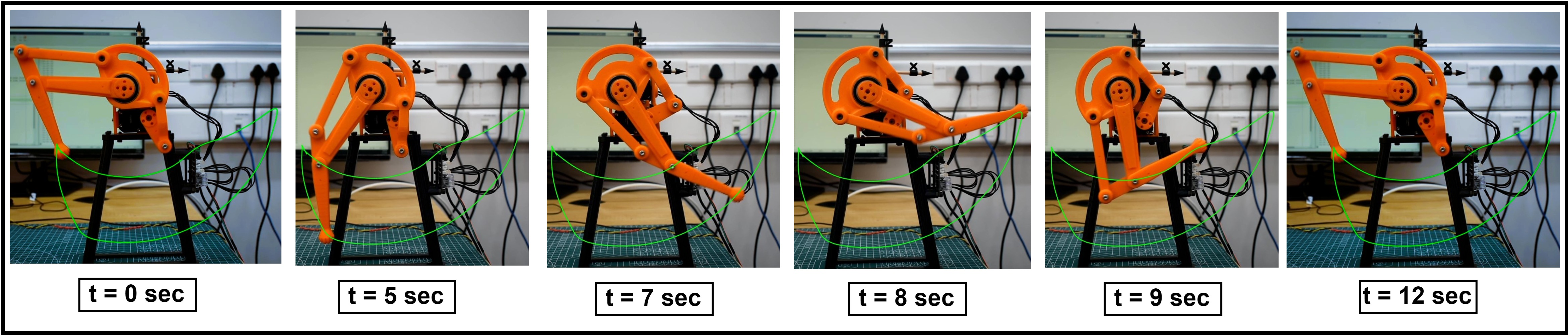}
    \caption{Time lapse of the workspace boundary generation on the test rig of the leg}
    \label{workspace_frames}
\end{figure}

The point cloud obtained earlier contains all valid task-space coordinates and corresponding joint-space values. To validate the workspace, all foot locations on the workspace boundary are selected for implementation on the hardware. For these foot locations, a joint-angle vector is provided to the actuators. As the reachable workspace has already been obtained through workspace analysis via MATLAB simulation, as shown in Fig. \ref{task_space}, a similar contour is observed in Fig. \ref{workspace_frames}. Additionally, hardware implementation of the workspace generation by the leg can be found at this \href{https://youtu.be/drWc4Ag9Xu4}{link}. Moreover, the overlay workspace contour across the different frames in Fig. \ref{workspace_frames} is created by manually tracking the leg's foot location in each frame and thus contains some noise, which may make it appear slightly different from the simulated one.
\subsection{Gait generation}
To implement the leg in the quadruped, a few gaits are planned to assess the leg's function during the swing and stance phases of the robot in quadruped mode. Again, this gait planning is based on fitting the desired trajectory to the available points in the previously obtained point cloud. Once the curve is fitted, the required trajectory is obtained in both the task and the joint space.
\paragraph{\textbf{Crouch gait:}} For this gait, basic ellipse geometry is selected. The reason for choosing this shape is primarily due to the following:
\begin{itemize}
    \item Ellipses are simple and smooth geometries and can be easily fitted to the point cloud.
    \item With an ellipse, it is easy to increase or decrease the length of the stance and swing phases of the gait by varying its major and minor axes.
    \item The center of the ellipse helps to locate the gait inside the workspace easily, to avoid reaching the less manipulable regions. 
\end{itemize}  
However, other gaits can also be planned using the available point cloud; to complete the foundational work for the leg, it is decided to stick with this geometry in the present work.\\ Now the equation of the ellipse used for the crouch gait can be written as:
\begin{equation}
    \begin{bmatrix}
        X(t)\\Z(t)
    \end{bmatrix}=\begin{bmatrix}
        x_{c}+a\cos(t)\\z_{c}+b\sin(t)
    \end{bmatrix}
    \label{ellipse}
\end{equation}
Where, $X(t)$ and $Z(t)$ are the coordinates of the elliptical trajectory in task space. $a$ and $b$ are the major and minor axes, respectively, of the ellipse. $x_{c}$ and $z_{c}$ are the coordinates of the center of the ellipse, which can be used to shift the trajectory within the workspace. Also, $t$ is used as the parameter, such that $0\leq t \leq 2\pi$, is indexed in 150 points, which, however, can be increased or decreased based on the smoothness of the trajectory required. 
\begin{figure}
    \centering
    \includegraphics[width=0.8\linewidth]{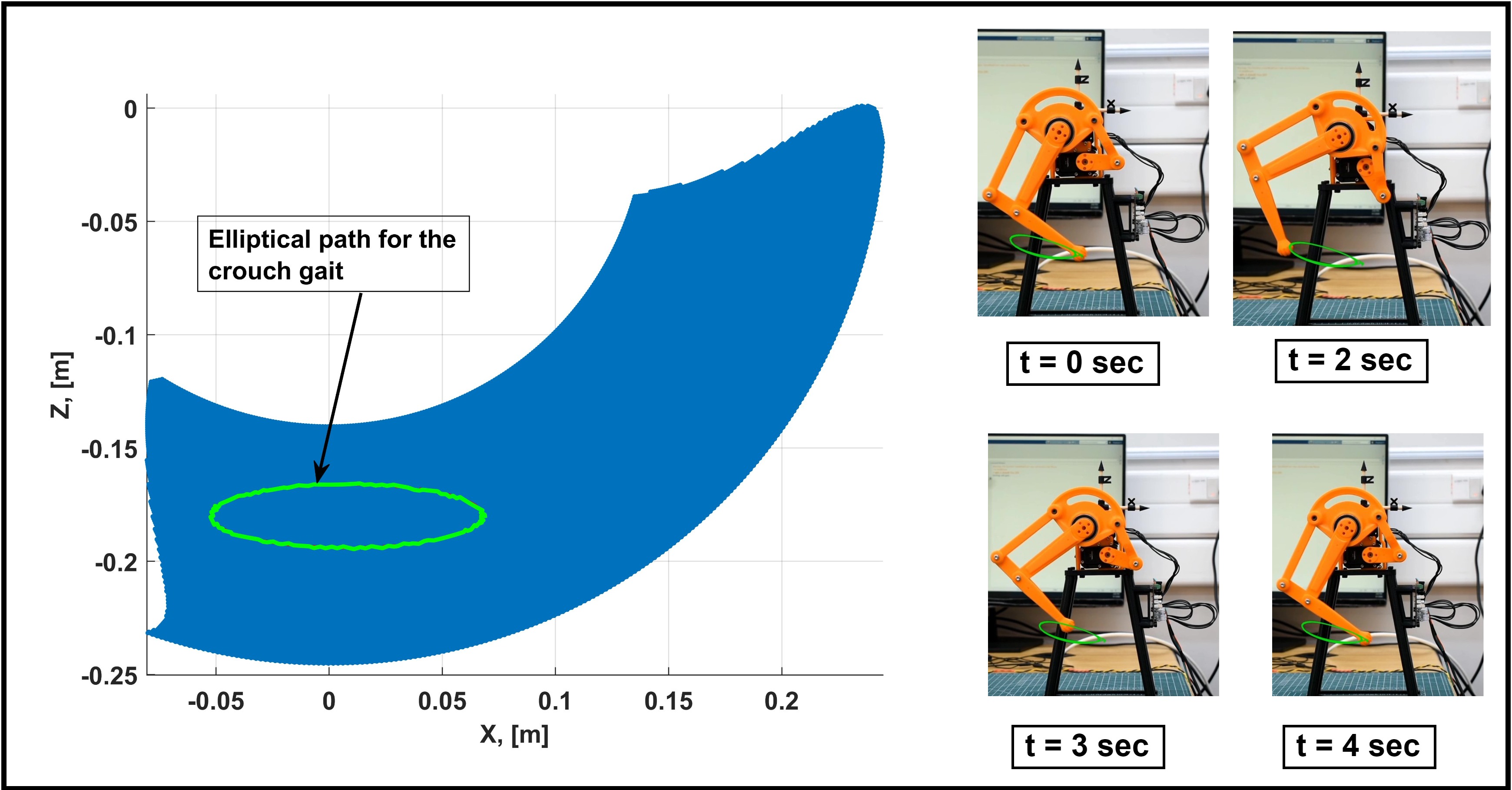}
    \caption{Planned crouch gait within the workspace (Left), with transition on test bench (Right)}
    \label{crouch_frames}
\end{figure}
\\For the crouch gait, the center of the ellipse was taken at, $x_{c}=0.008\,\text{m}$  and $z_{c}=-0.18\,\text{m}$, with $a=0.06\,\text{m}$ and $b=0.014\,\text{m}$. Now, based on the desired trajectory's elliptical path parameters, an algorithm was developed that finds the closest point to the previously obtained point using MATLAB's built-in PCHIP (Piecewise cubic Hermite interpolation) function to obtain a smooth trajectory from discrete workspace points. The final smooth output trajectory, overlaid onto the workspace, as shown in Fig. \ref{crouch_frames}, along with transient frames from the test bench implementation. The experimental implementation of this gait can be found at this \href{https://youtu.be/3Y_RwLkK6BU}{link}.
\paragraph{\textbf{Trot gait:}} For the trot gait of the quadruped, the same elliptical trajectory was used as for the crouch gait, but with different geometrical parameters. Here, $a=0.08\,\text{m}$, $b=0.035\,\text{m}$ and $x_{c}=0.02\,\text{m}$, $z_{c}=-0.21$ is used. 
\begin{figure}
    \centering
    \includegraphics[width=0.8\linewidth]{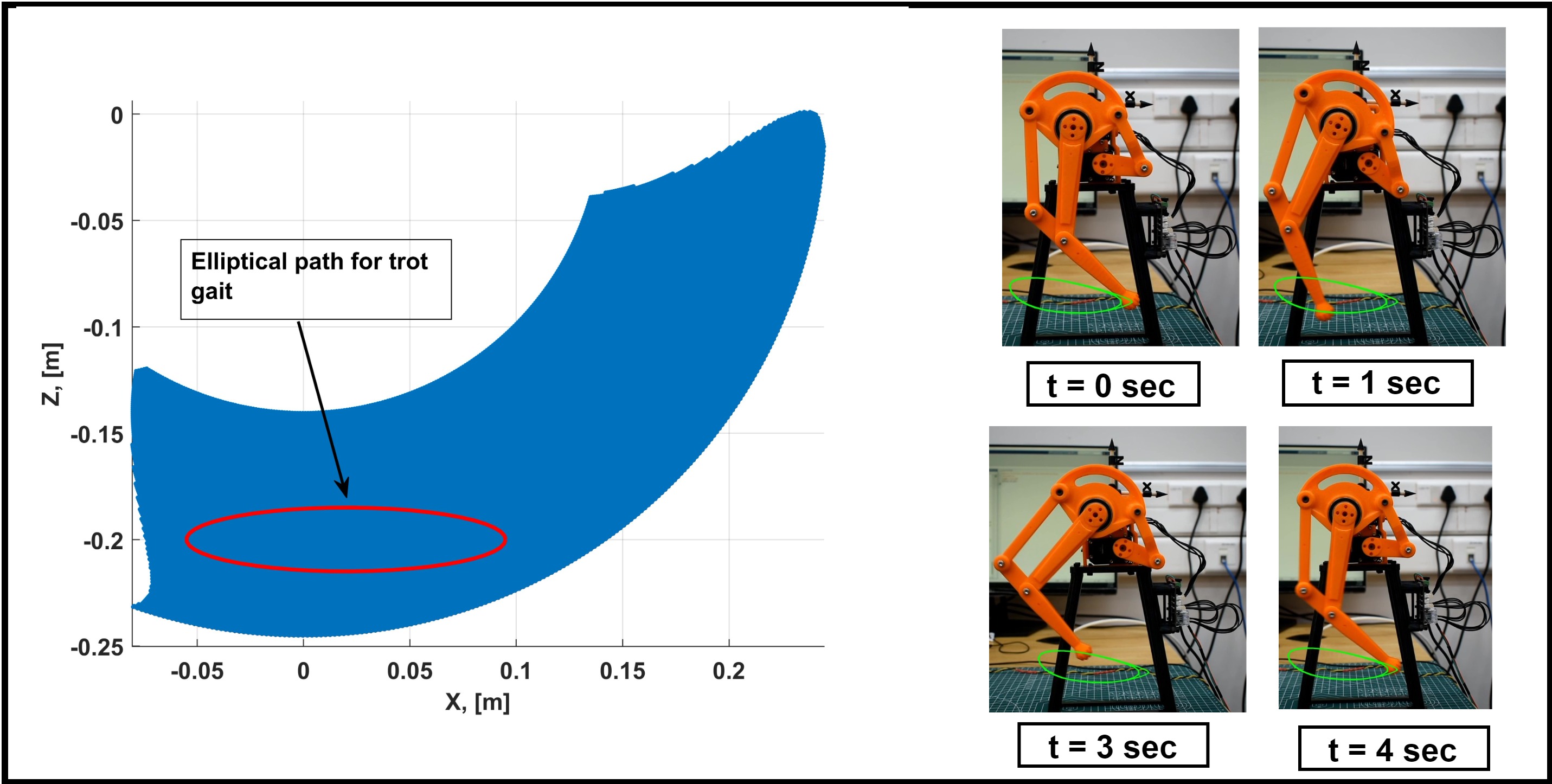}
    \caption{Planned trot gait overlaid onto the workspace (left), with temporal frame-wise representation onto the test bench (right)}
    \label{trot_frames}
\end{figure}
In this case, the center of the ellipse is shifted downward on the negative of the Z-axis to stretch the leg a little bit more compared to what it is in the crouched gait. Also, the trot gait's stride length is set slightly longer than the crouch to make it more appropriate for normal walking. The hardware implementation of this gait can be compared with Fig.\ref{trot_frames} and a video demonstration can be seen at this \href{https://youtu.be/uYhXusMkesA}{link}. Moreover, from the same figure, it can be argued that the center of the ellipse can be shifted a little more downward, which is true, but it should be done with the fact in mind that, in the lower region of the workspace, $\sigma_{min}$ of the $\mathbf{J_{leg}}$ is reducing, due to which actuator motion was found to be jittery.  
\paragraph{\textbf{Squat gait:}} The principal objective of this gait's creation is to facilitate the transition of the quadruped from a fully seated position to an upright stance at a designated place. This can be done by flexing the thigh and the knee link of the leg and corresponding joint angles $\theta_{2}$ and $\theta_{3}$. 
\begin{figure}
    \centering
    \includegraphics[width=0.75\linewidth]{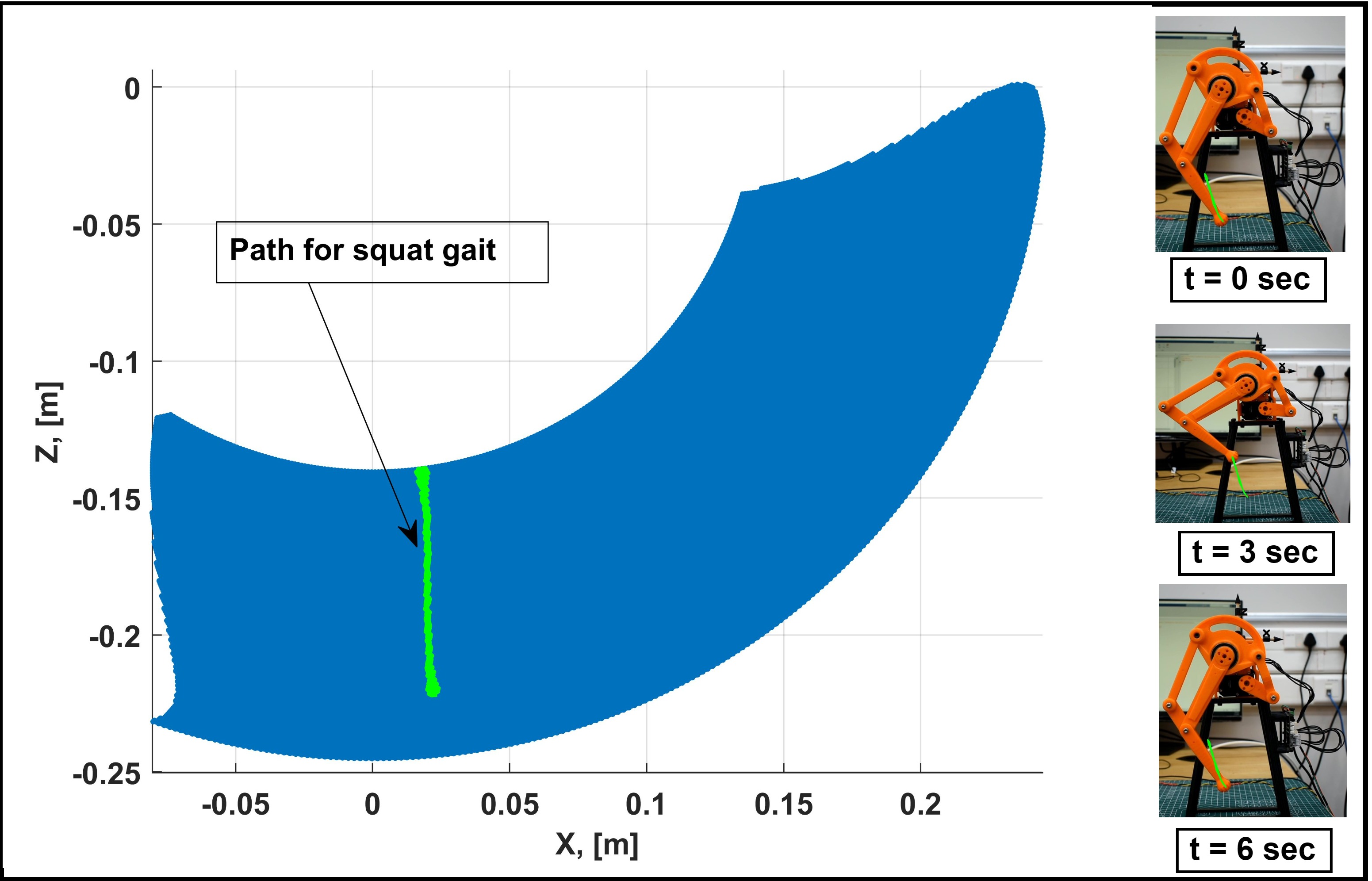}
    \caption{Planned squat gait, overlaid onto the workspace (left), with time lapse of squat gait on hardware (right)}
    \label{squat_gait}
\end{figure}
For the implementation of this motion, only a straight path was required, keeping $X(t)$ as constant and varying $Z(t)$ only. But in such a straight-line trajectory, the actuator motion will be jerky due to the sudden jumps in the velocity; therefore, it is decided to make the trajectory with small perturbations in $X(t)$ and parametrize the $Z(t)$ with a sinusoidal function. This trajectory can be described as follows:
\begin{equation}
\begin{bmatrix}
X(t)\\Z(t)
\end{bmatrix} =
\begin{bmatrix}
x_{c}+0.002\cos^{3}(t)\\
z_{\min}+\left(z_{\max}-z_{\min}\right)
\left(\dfrac{1-\cos(t)}{2}\right)
\end{bmatrix}
\label{squat}
\end{equation}
where $X(t)$ and $Z(t)$ represent the horizontal and vertical coordinates of the foot trajectory, respectively. The parameter $x_c=0.02\,\text{m}$ denotes the horizontal center position of the trajectory, while $z_{\min}=-0.22\,\text{m}$ and $z_{\max}=-0.14\,\text{m}$ correspond to the minimum and maximum vertical positions of the squat motion. The term $0.002\cos^{3}(t)$ introduces a tiny lateral perturbation to improve numerical robustness during local workspace projection. The cosine-based parameterization of $Z(t)$ ensures smooth velocity transitions and minimizes actuator jerk during motion initiation and termination. The implemented trajectory in the MATLAB simulation and hardware implementation can be seen in Fig. \ref{squat_gait}, along with a video demonstration provided at this \href{https://youtu.be/vlEb8t99mdk}{link}.
\subsection{Reconfiguration of the locomotor module}
In the overall proposed system, the leg is used only for terrestrial locomotion; however, for aerial locomotion, it needs to be reconfigured appropriately to meet the requirements of the aerial vehicle. Therefore, to meet this requirement, this leg is equipped with a $90\,\text{mm}$ EDF (Electric Ducted Fan) module. This helps to generate the required thrust for the system to be airborne. But, for that, the leg needs to be reconfigured from the leg pose to the aerial pose. In this section, the transition motion is studied, which will be very useful later in the work for switching modes.
\begin{figure}
    \centering
    \includegraphics[width=1\linewidth]{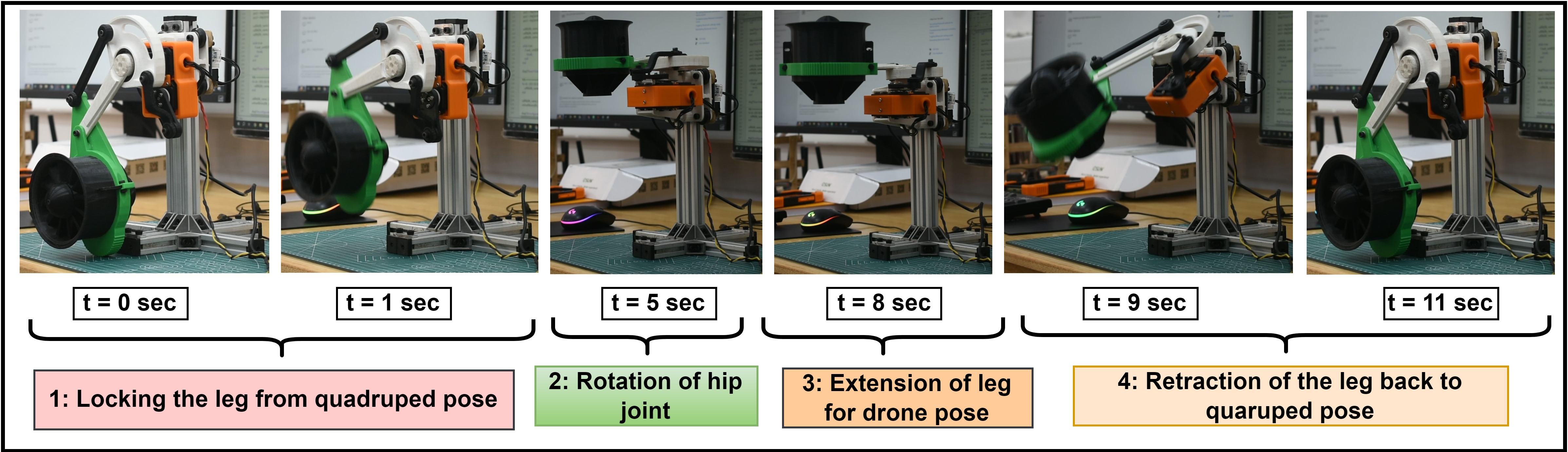}
    \caption{Time lapse of reconfiguration of the locomotor}
    \label{reconfig_leg}
\end{figure}
The reconfiguration of the leg is planned in four discrete phases, as shown in Fig. \ref{reconfig_leg}. These four phases can be described as follows:
\begin{itemize}
    \item In phase one, the leg is locked to a safe pose with hip joint angle $\theta_{1}=0^\circ$, thigh joint angle $\theta_{2}=180^\circ$, and knee joint angle $\theta_{3}=114^\circ$.
    \item In phase two, the thigh and knee joint angles are kept constant, and the hip joint is flexed to $90^\circ$.
    \item In phase three, hip joint is locket at $\theta_{1}=90^\circ$, thigh and knee joints are flexed to $\theta_{2}=270^\circ$ and $\theta_{3}=29^\circ$ respectively.
    \item In phase four, all three actuators are unloaded simultaneously, using $\theta_{1}=0^\circ$, $\theta_{2}=235^\circ$, and $\theta_{3}=29^\circ$.
\end{itemize}
The sequence of all four phases can also be verified from a real hardware-implementation video provided at this \href{https://youtu.be/TuV2PadmSyU}{link}.
\section{Conclusion and future scope} In this work, the kinematic analysis of a 3-degree-of-freedom, double, four-bar closed-loop-based single leg is discussed with experimental validation. This leg, in terms of the original proposed work, is named "locomotor," which can work as a leg during quadruped mode and a rotor arm during the aerial mode. This leg is crucial for transitioning between the two operational modes; thus, it is analyzed through singularity analysis, leading to the formulation of its Jacobian, which is instrumental in designing the gaits necessary for the terrestrial mode and mid-air aerial reconfiguration in the aerial mode. The purpose of this work is to lay a solid foundation for the overall system. Therefore, it is very important that all aspects of the locomotor module be thoroughly explored and tested in various ways to avoid kinematic design-related problems in the future. Since the work presented here in this article was the first phase of the original full system. In the next work, a full dynamic analysis of the leg with a working EDF rotor, ground interaction, and contact mechanics will be explored with implementation on a real-world robot.
\section*{Declaration of competing interest}
The authors declare that they have no known competing financial
interests or personal relationships that could have appeared to
influence the work reported in this paper.
\section*{Acknowledgement}
This research is supported by the IIT Palakkad Technology IHub Foundation [Grant no: IPTIF/IP/09] and Anusandhan National Research Foundation [Grant no: CRG/2023/004646].
\section*{Declaration of Generative AI} Authors would like to declare that the use of licensed Quillbot software is used to improve grammar, flow, and readability of the text. However, authors have reviewed and edited the content as needed and take full responsibility for the final content and accuracy of the work.
\bibliographystyle{unsrt}
\bibliography{ref.bib}
\appendix
\section*{Appendix}
\addcontentsline{toc}{section}{Appendix}
\section{constants used in the derivation of Closed-Loop Kinematics}
\subsection{Loop: AEFGA}
$k_{1} = AE\cos{a_{1}}$\\
$k_{2} = AG\cos{\theta_{2}}$\\
$k_{3} = AE\sin{a_{1}}$\\
$k_{4} = AG\sin{a_{4}}$\\
$k_{5} = FG^{2} + k_{1}^{2} + k_{2}^{2} + k_{3}^{2} + k_{4}^{2} - EF^{2} - 2k_{2}k_{1} - 2k_{3}k_{4}$\\
$k_{6} = 2GF(k_{2}-k_{1})$\\
$k_{7} = 2GF(k_{4}-k_{3})$\\
$A_{1} = k_{5}-k_{6}$\\
$B_{1} = 2k_{7}$\\
$C_{1} = k_{5}+k_{6}$\\
$D_{1} = B_{1}^2-4A_{1}C_{1}$\\
====================================\\
$k_{8} = k_{1}^{2}+k_{2}^{2}+k_{3}^{2}+k_{4}^{2}+EF^{2}-GF^{2}-2k_{1}k_{2}-2k_{3}k_{4}$\\
$k_{9} = 2EF(k_{1}-k_{2})$\\
$k_{10} = 2EF(k_{3}-k_{4})$\\
$A_{2} = k_{8}-k_{9}$\\
$B_{2} = 2k_{10}$\\
$C_{2} = k_{8}+k_{9}$\\
$D_{2} = B_{2}^{2}-4A_{2}C_{2}$
\label{A11}
\subsection{Loop: ABCDE}
$k_{11} = AB\cos{a_{5}}$\\
$k_{12} = BC\cos{a_{6}}$\\
$k_{13} = AB\sin{a_{5}}$\\
$k_{14} = BC\sin{a_{6}}$\\
$k_{15} = k_{11}^{2}+k_{12}^{2}+k_{13}^{2}+k_{14}^{2}+DC^{2}-AD^{2}+2K_{11}k_{12}+2k_{13}k_{14}$\\
$k_{16} = 2DC(k_{11}+k_{12})$\\
$k_{17}  =2DC(K_{13}+k_{14})$\\
$A_{3} = k_{15}+k_{16}$\\
$B_{3} = -2k_{17}$\\
$C_{3} = k_{15}-k_{16}$\\
$D_{3} = B_{3}^{2}-4A_{3}C_{3}$\\
====================================\\
$k_{18} = k_{11}^{2}+k_{12}^{2}+k_{13}^{2}+k_{14}^{2}+AD^{2}-DC^{2}+2k_{11}k_{12}+2k_{13}k_{14}$\\
$k_{19} = 2AD(k_{11}+k_{12})$\\
$k_{20} = 2AD(k_{13}+k_{14})$\\
$A_{4} = k_{18}+k_{19}$\\
$B_{4} = -2k_{20}$\\
$C_{4} = k_{18}-k_{19}$\\
$D_{4} = B_{4}^{2}-4A_{4}C_{4}$
\label{A12}
\section{Known geometrical parameters of the two closed-loop chains of the rig}
Angles $\theta_{2}$ and $\theta_{3}$ are already known from the inverse kinematics, and are given in equations \eqref{eq_12} and \eqref{eq_14}.\\
$a_{3} = (\theta_{2}-\pi+\theta_{3})$\\
$a_{5} = 307.35^\circ$\\
$\gamma = 144.75^\circ$\\
$a_{1} = a_{8}+\gamma$\\
AB = 0.0478 m\\
BC = 0.04 m\\
DC = 0.05787 m\\
AD = 0.045 m\\
AE = 0.045 m\\
EF = 0.098 m\\
FG = 0.04 m\\
AG = 0.11 m\\
\label{A2}
\section{Expression for the parent Chain Jacobian $\mathbf{J_{parent}}$}
\begin{equation*}
\mathbf{J}_{parent} =
\begin{bmatrix}
\sin\psi\sin\theta\,\sigma_4 - \cos\theta\,\sigma_2 &
\sin(\phi+\theta_1)\sin\theta\,\sigma_8 - \sigma_9\sigma_6 &
s_2\sin(\phi+\theta_1)\sin\theta\,\sigma_{13} - s_2\sigma_{11}\sigma_9 \\

-\cos\theta\,\sigma_3 - \cos\psi\sin\theta\,\sigma_4 &
\sin(\phi+\theta_1)\sin\theta\,\sigma_7 - \sigma_{10}\sigma_6 &
s_2\sin(\phi+\theta_1)\sin\theta\,\sigma_{12} - s_2\sigma_{11}\sigma_{10} \\

\sin\psi\sin\theta\,\sigma_3 + \cos\psi\sin\theta\,\sigma_2 &
\sigma_{10}\sigma_8 - \sigma_9\sigma_7 &
s_2\sigma_{13}\sigma_{10} - s_2\sigma_{12}\sigma_9 \\

\cos\psi\sin\theta & \sigma_{10} & \sigma_{10} \\
\sin\psi\sin\theta & \sigma_5 & \sigma_5 \\
\cos\theta & \sigma_1 & \sigma_1
\end{bmatrix}
\end{equation*}
where:
\begin{equation*}
\left.
\begin{aligned}
\sigma_1 &= -\sin(\phi+\theta_1)\sin\theta \\
\sigma_2 &= s_1\sigma_{16} + s_2\sigma_{13} - h_2\sigma_9 \\
\sigma_3 &= s_1\sigma_{15} + s_2\sigma_{12} - h_2\sigma_{10} \\
\sigma_4 &= s_1\sigma_{14} + s_2\sigma_{11} - h_2\sin(\phi+\theta_1)\sin\theta \\
\sigma_5 &= \sin\theta_1\,\sigma_{22} - \cos\theta_1\,\sigma_{21} \\
\sigma_6 &= s_1\sigma_{14} + s_2\sigma_{11} \\
\sigma_7 &= s_1\sigma_{15} + s_2\sigma_{12} \\
\sigma_8 &= s_1\sigma_{16} + s_2\sigma_{13} \\
\sigma_9  &= \cos\theta_1\,\sigma_{21} - \sin\theta_1\,\sigma_{22} \\
\sigma_{10} &= \cos\theta_1\,\sigma_{20} - \sin\theta_1\,\sigma_{19} \\
\sigma_{11} &= \cos\theta_3\,\sigma_{14} + \sin\theta_3\big(\cos\theta\cos\theta_2 + \cos(\phi+\theta_1)\sin\theta\sin\theta_2\big) \\
\sigma_{12} &= \cos\theta_3\,\sigma_{15} - \sin\theta_3\big(\sin\theta_2\,\sigma_{17} + \cos\psi\cos\theta_2\sin\theta\big) \\
\sigma_{13} &= \cos\theta_3\,\sigma_{16} - \sin\theta_3\big(\sin\theta_2\,\sigma_{18} - \cos\theta_2\sin\psi\sin\theta\big) \\
\sigma_{14} &= \cos\theta\sin\theta_2 - \cos(\phi+\theta_1)\cos\theta_2\sin\theta \\
\sigma_{15} &= \cos\theta_2\,\sigma_{17} - \cos\psi\sin\theta\sin\theta_2 \\
\sigma_{16} &= \cos\theta_2\,\sigma_{18} + \sin\psi\sin\theta\sin\theta_2 \\
\sigma_{17} &= \cos\theta_1\,\sigma_{19} + \sin\theta_1\,\sigma_{20} \\
\sigma_{18} &= \cos\theta_1\,\sigma_{22} + \sin\theta_1\,\sigma_{21} \\
\sigma_{19} &= \sin\phi\sin\psi - \cos\phi\cos\psi\cos\theta \\
\sigma_{20} &= \cos\phi\sin\psi + \cos\psi\cos\theta\sin\phi \\
\sigma_{21} &= \cos\phi\cos\psi - \cos\theta\sin\phi\sin\psi \\
\sigma_{22} &= \cos\psi\sin\phi + \cos\phi\cos\theta\sin\psi
\end{aligned}
\right\}
\quad \text{auxiliary terms}
\end{equation*}
\label{A3}
\section{Expression for the foot position of the leg with respect to the inertial frame \{FR4\}, using forward kinematics} 
\begin{equation*}
\begin{aligned}
c_{\psi} &= \cos\psi,
& s_{\psi} &= \sin\psi,
\\
c_{\phi} &= \cos\phi,
& s_{\phi} &= \sin\phi,
\\
c_{\theta} &= \cos\theta,
& s_{\theta} &= \sin\theta,
\\
c_{11} &= \cos\theta_{11},
& s_{11} &= \sin\theta_{11},
\\
c_{12} &= \cos\theta_{12},
& s_{12} &= \sin\theta_{12},
\\
c_{13} &= \cos\theta_{13},
& s_{13} &= \sin\theta_{13}
\end{aligned}
\end{equation*}
\begin{equation*}
\left.
\begin{aligned}
A
&=
c_{11}
\left(
s_{\phi}s_{\psi}
-
c_{\psi}s_{\theta}c_{\phi}
\right)
+
s_{11}
\left(
s_{\psi}c_{\phi}
+
c_{\psi}s_{\phi}s_{\theta}
\right)
\\[1ex]
B
&=
-c_{11}
\left(
s_{\psi}c_{\phi}
+
c_{\psi}s_{\phi}s_{\theta}
\right)
+
s_{11}
\left(
s_{\phi}s_{\psi}
-
c_{\psi}s_{\theta}c_{\phi}
\right)
\\[1ex]
C
&=
s_{\phi}c_{\psi}
+
s_{\psi}s_{\theta}c_{\phi}
\\[1ex]
D
&=
s_{\psi}s_{\phi}
-
c_{\psi}s_{\theta}c_{\phi}
\\[1ex]
E
&=
c_{11}c_{\theta}c_{\phi}
-
s_{11}c_{\theta}s_{\phi}
\end{aligned}
\right\}
\end{equation*}
\begin{equation*}
{}^{I}\mathbf{P}_{FR4}
=
\begin{bmatrix}
P_{x}\\
P_{y}\\
P_{z}
\end{bmatrix}
\end{equation*}
\begin{equation*}
\left.
\begin{aligned}
P_{x}
&=
x
-s_2
\Big[
c_{13}
\left(
c_{12}A
-
c_{\psi}c_{\theta}s_{12}
\right)
-
s_{13}
\left(
s_{12}A
+
c_{\psi}c_{12}c_{\theta}
\right)
\Big]
\\
&\quad
+\frac{w}{2}D
-s_1
\left(
c_{12}A
-
c_{\psi}c_{\theta}s_{12}
\right)
-h_2B
+h_1c_{\psi}c_{\theta}
+\frac{l}{2}c_{\psi}c_{\theta}
\\[2ex]
P_{y}
&=
y
+ s_2
\Big[
s_{13}
\left(
c_{12}C
+
s_{\psi}c_{\theta}s_{12}
\right)
+
c_{13}
\left(
s_{12}C
-
s_{\psi}c_{12}c_{\theta}
\right)
\Big]
\\
&\quad
+\frac{w}{2}
\left(
c_{\psi}c_{\phi}
+
s_{\psi}s_{\theta}s_{\phi}
\right)
\\
&\quad
+s_1
\left(
c_{12}C
+
s_{\psi}c_{\theta}s_{12}
\right)
+h_2
\left(
c_{11}
(c_{\psi}c_{\phi}
+
s_{\psi}s_{\theta}s_{\phi})
+
s_{11}C
\right)
\\
&\quad
+h_1s_{\psi}c_{\theta}
+\frac{l}{2}s_{\psi}c_{\theta}
\\
P_{z}
&=
z
+s_1
\left(
-c_{\theta}s_{12}
+
c_{12}E
\right)
\\
&\quad
-h_2
\left(
c_{11}c_{\theta}s_{\phi}
+
s_{11}c_{\theta}c_{\phi}
\right)
\\
&\quad
-s_2
\Big[
s_{13}
\left(
-c_{\theta}c_{12}
-
s_{12}E
\right)
+
c_{13}
\left(
-c_{\theta}s_{12}
+
c_{12}E
\right)
\Big]
\\
&\quad
-h_1s_{\theta}
-\frac{l}{2}s_{\theta}
-\frac{w}{2}c_{\theta}s_{\phi}
\end{aligned}
\right\}
\end{equation*}
\label{forward_kinematics}
\end{document}